\def\deanonymized{}
\documentclass{article} 
\usepackage{iclr2027_conference,times}

\usepackage{amsmath,amssymb}

\usepackage{amsmath,amsfonts,bm}

\def\eqref#1{equation~\ref{#1}}

\def\1{\bm{1}}

\DeclareMathAlphabet{\mathsfit}{\encodingdefault}{\sfdefault}{m}{sl}
\SetMathAlphabet{\mathsfit}{bold}{\encodingdefault}{\sfdefault}{bx}{n}

\newcommand{\ScaleCeiling}{0.99997}
\newcommand{\ScaleRandom}{0.5024}
\newcommand{\ScaleInterval}{0.49755}
\newcommand{\CeilMin}{0.968}
\newcommand{\CeilMax}{1.000}
\newcommand{\CeilMedian}{0.997}
\newcommand{\CeilMaxExact}{1.000000}

\newcommand{\CompRangeKRounded}{0.005}
\newcommand{\CompSigmaMax}{0.1444}

\newcommand{\CompEffectKFive}{0.0545}
\newcommand{\CompSepTypical}{0.146}
\newcommand{\CovRsqPooled}{0.101}

\newcommand{\PctEnSubsets}{71.8}

\newcommand{\PctEnMerges}{75.2}

\newcommand{\PctOthersLow}{38}
\newcommand{\PctOthersHigh}{47}

\newcommand{\GranCommon}{+0.0768}
\newcommand{\GranDifferential}{0.0564}
\newcommand{\GranRho}{-0.773}
\newcommand{\GranRhoP}{3.7e-06}
\newcommand{\GranWeakSix}{+0.1225}
\newcommand{\GranStrongSix}{+0.0284}

\newcommand{\GranHubertRandomFine}{0.317}
\newcommand{\GranHubertRandomCoarse}{0.224}

\newcommand{\HierModelsThr}{+0.922}
\newcommand{\HierModelsMet}{+0.929}
\newcommand{\HierModelsThrCI}{[+0.59, +1.00]}
\newcommand{\HierModelsMetCI}{[+0.80, +0.97]}
\newcommand{\HierModelsThrN}{8}
\newcommand{\HierModelsMetN}{27}
\newcommand{\ThrDifferential}{0.0309}
\newcommand{\ThrMin}{0}
\newcommand{\ThrMax}{100}

\newcommand{\ThrPlBefore}{-0.047}
\newcommand{\ThrPlAfter}{-0.002}

\newcommand{\DiagRuRank}{fourth}

\newcommand{\DiagCorr}{+0.118}
\newcommand{\DiagCorrP}{0.676}

\newcommand{\MetDifferential}{0.0361}
\newcommand{\MetGapEuc}{0.0467}

\newcommand{\MetRsaFactor}{2.5}
\newcommand{\MetCkaSlope}{+0.00722}
\newcommand{\AggDifferentialPct}{2.33\%}

\newcommand{\QuantGridEs}{1.21e-02}

\newcommand{\QuantTied}{Four}

\newcommand{\LayerSurviving}{0}
\newcommand{\LayerPairs}{32}

\newcommand{\LayerInflationPct}{5.2\%}
\newcommand{\LayerGap}{0.0564}
\newcommand{\LayerShare}{46\%}
\newcommand{\CloseCompIqrShift}{+0.0533}
\newcommand{\CloseCompIqrPct}{+10.7\%}
\newcommand{\CloseCompIqrPctCI}{[+9.5, +12.1]}
\newcommand{\CloseCompSigmaShift}{+0.0435}
\newcommand{\CloseCompSigmaPct}{+8.7\%}
\newcommand{\CloseCompSigmaPctCI}{[+8.1, +9.3]}
\newcommand{\CloseThrShift}{+0.0428}
\newcommand{\CloseThrPct}{+8.6\%}
\newcommand{\CloseThrPctCI}{[-4.1, +21.1]}
\newcommand{\CloseGranShift}{+0.0411}
\newcommand{\CloseGranPct}{+8.3\%}
\newcommand{\CloseGranPctCI}{[-7.9, +16.6]}
\newcommand{\CloseDistShift}{+0.0358}
\newcommand{\CloseDistPct}{+7.2\%}
\newcommand{\CloseDistPctCI}{[-5.4, +16.5]}
\newcommand{\CloseLayerShift}{-0.0242}
\newcommand{\CloseLayerPct}{-4.9\%}
\newcommand{\CloseLayerPctCI}{[-12.0, -0.8]}
\newcommand{\CloseRankShift}{-0.0210}
\newcommand{\CloseRankPct}{-4.2\%}
\newcommand{\CloseRankPctCI}{[-18.5, +8.5]}
\newcommand{\CloseAggShift}{+0.0039}
\newcommand{\CloseAggPct}{+0.8\%}
\newcommand{\CloseAggPctCI}{[-2.5, +4.0]}
\newcommand{\CloseCompIqrZfullPct}{+12.7\%}
\newcommand{\RankDrawsKFourteen}{+0.926}
\newcommand{\RankDrawsKFourteenCI}{[+0.921, +0.930]}
\newcommand{\RankDrawsKFourteenTrained}{+0.838}
\newcommand{\RankDrawsKFourteenTrainedCI}{[+0.832, +0.847]}
\newcommand{\RankDrawsKFive}{+0.539}
\newcommand{\RankDrawsKFiveCI}{[+0.512, +0.565]}
\newcommand{\RankDrawsKFiveTrained}{+0.313}
\newcommand{\RankDrawsKFiveTrainedCI}{[+0.282, +0.344]}
\newcommand{\RankDrawsKTwenty}{+0.969}
\newcommand{\RankDrawsKTwentyCI}{[+0.965, +0.972]}
\newcommand{\RankDrawsKTwentyTrained}{+0.921}
\newcommand{\RankDrawsKTwentyTrainedCI}{[+0.915, +0.924]}
\newcommand{\RankDrawsKFiveQuartile}{+0.06}

\newcommand{\MetAgreeRsaCkaAll}{+0.969}
\newcommand{\MetAgreeRsaCkaAllCI}{[+0.913, +0.985]}
\newcommand{\MetAgreeRsaProcAll}{+0.967}
\newcommand{\MetAgreeRsaProcAllCI}{[+0.894, +0.988]}
\newcommand{\MetAgreeCkaProcAll}{+0.944}
\newcommand{\MetAgreeCkaProcAllCI}{[+0.893, +0.974]}
\newcommand{\MetAgreeRsaCkaTrained}{+0.868}
\newcommand{\MetAgreeRsaCkaTrainedCI}{[+0.617, +0.959]}
\newcommand{\MetAgreeRsaProcTrained}{+0.909}
\newcommand{\MetAgreeRsaProcTrainedCI}{[+0.638, +0.976]}
\newcommand{\MetAgreeCkaProcTrained}{+0.779}
\newcommand{\MetAgreeCkaProcTrainedCI}{[+0.609, +0.918]}
\newcommand{\MetAgreeRsaProcEnAll}{+0.872}
\newcommand{\MetAgreeRsaCkaEnTrained}{+0.856}
\newcommand{\MetAgreeRsaProcEnTrained}{+0.635}
\newcommand{\MetAgreeCkaProcEnTrained}{+0.785}

\newcommand{\MetCeilRsa}{\ScaleCeiling}
\newcommand{\MetFloorRsa}{\ScaleRandom}
\newcommand{\MetIntervalRsa}{\ScaleInterval}
\newcommand{\MetGapRsa}{\MetGapEuc}
\newcommand{\MetGapRsaPct}{9.4\%}
\newcommand{\MetGapRsaPctCI}{[4.8, 21.4]}
\newcommand{\MetCeilCka}{1.0000}
\newcommand{\MetFloorCka}{0.6669}
\newcommand{\MetIntervalCka}{0.3331}
\newcommand{\MetGapCka}{+0.0289}
\newcommand{\MetGapCkaPct}{8.7\%}
\newcommand{\MetGapCkaPctCI}{[5.6, 18.2]}
\newcommand{\MetCeilProc}{1.0000}
\newcommand{\MetFloorProc}{0.8258}
\newcommand{\MetIntervalProc}{0.1742}
\newcommand{\MetGapProc}{+0.0117}
\newcommand{\MetGapProcPct}{6.7\%}
\newcommand{\MetGapProcPctCI}{[4.1, 16.8]}
\newcommand{\VisPairs}{twelve}
\newcommand{\VisContestedArgmax}{one}
\newcommand{\VisContestedMatched}{five}
\newcommand{\VisContestedLoo}{three}

\newcommand{\CtrlArchitectures}{three}
\newcommand{\CtrlSeeds}{0, 1 and 2}
\newcommand{\CtrlInitStd}{0.02}
\newcommand{\CtrlTransformers}{4.57.6}
\newcommand{\CtrlTorch}{2.6.0}
\newcommand{\VitSampleRate}{16,000}
\newcommand{\VitMels}{128}
\newcommand{\VitFft}{2,048}
\newcommand{\VitMelHopMs}{32}
\newcommand{\VitWindowMs}{480}
\newcommand{\VitImage}{224}
\newcommand{\VitPatch}{16}
\newcommand{\VitLayers}{13}
\newcommand{\VitWidth}{768}
\newcommand{\VitNamedTen}{10}
\newcommand{\VitHopTenMs}{96}
\newcommand{\VitRateTen}{10.4}
\newcommand{\VitNamedFifty}{50}
\newcommand{\VitRateFifty}{31.25}
\newcommand{\ProvHubCheckpoints}{eleven}
\newcommand{\ProvAmbiguous}{six}
\newcommand{\PrecCells}{5,985}
\newcommand{\PrecMoved}{698}

\newcommand{\PrecMaxShiftPct}{19.5\%}
\newcommand{\PrecMaxBefore}{0.794}
\newcommand{\PrecMaxAfter}{0.891}
\newcommand{\PrecArgmaxMoved}{one}
\newcommand{\PrecDefaultMax}{0.0066}
\newcommand{\PrecAnchorBefore}{0.3280}
\newcommand{\PrecAnchorAfter}{0.3298}

\usepackage{booktabs}
\usepackage{graphicx}  
\usepackage{longtable}  
\usepackage{array}
\usepackage{makecell}  
\usepackage[safe]{tipa}  
\usepackage{microtype}
\usepackage{hyperref}
\usepackage{url}

\title{A Better Metric Will Not Save You:\\
What You Compare Beats How You Compare It}

\ifdefined\deanonymized
\author{Noureddine Khaous \& Guillaume Wisniewski \\
Universit\'e Paris Cit\'e, LLF, CNRS \\
75\,013 Paris, France \\
\texttt{noureddine.khaous@etu.u-paris.fr}, \texttt{guillaume.wisniewski@u-pariscite.fr}}
\iclrfinalcopy
\else
\author{Anonymous authors}
\fi

\begin{document}

\maketitle
\ifdefined\deanonymized\lhead{Preprint}\fi

\begin{abstract}
Representational similarity measures (RSA, CKA, Procrustes) are chosen for the
transformations they treat as irrelevant, and all of them take the set of
conditions as given. That set is fixed by two decisions, which inputs are
admitted and how they are grouped, both taken before the matrix a measure
operates on exists: no invariance property reaches them. We measure what they
are worth where the arrangement the conditions ought to have is known
independently of the systems compared, the acoustic organisation of vowels,
across fifteen languages and twenty-seven representations. Changing which
categories are compared, at a fixed number of them, moves the gap between two
representations by more than twice the effect being measured, while changing how
many moves it not at all. In this setting rankings of models are identified
under these choices, and rankings across languages are not.
\end{abstract}

\section{Introduction}
\label{sec:intro}

Claims of the form \emph{model A represents X more like B than C does} are now
routine: between successive layers of a network
\citep{raghu2017,kornblith2019}, between networks differing only in random
initialisation \citep{ding2021}, and between artificial models and biological
recordings \citep{kriegeskorte2008,khalighrazavi2014}, among others. They are
used to decide which architectures are worth pursuing and which accounts of a
cognitive function remain viable.

Comparing representations is harder than it looks, because it is not obvious
what \emph{similar representation} could mean. Two systems given the same inputs
produce representations of different widths and in different bases, with no
correspondence between individual dimensions: a dimension of a transformer's
hidden state has no counterpart in a second network, still less in a cortical
recording. No shared coordinate system exists, so there is nothing to correlate,
and were a number nonetheless produced, nothing would say whether it is a large
one. Nor is either representation interpretable: with no account of what any
dimension encodes, we cannot even specify what the two systems ought to agree
about. Similarity must be defined without knowing what is being compared.

The measures that answer this, such as Representational Similarity Analysis
\citep{kriegeskorte2008}, Centred Kernel Alignment \citep{kornblith2019} or
Procrustes-based shape metrics \citep{williams2021}, were introduced in
cognitive neuroscience, and machine learning adopted them largely as they stood.
They compare the \emph{organisation} of each representational space rather than
the representations themselves: one fixes a set of $N$ \emph{conditions} and
computes, within each system separately, the $N \times N$ matrix of pairwise
distances between its representations; the two matrices are then compared.
Fixing that set is two decisions, and we keep them apart throughout:
\emph{which inputs are admitted} to the comparison at all, and \emph{how the
admitted inputs are grouped} into the units among which distances are taken. A
condition is thereby characterised by its relations to the other conditions and
never by the dimensions encoding it, which is what dissolves the correspondence
problem.

There is more than one such measure, however, and applied to the same data they
disagree \citep{ding2021}. A substantial literature has accordingly grown up
around choosing among them, proceeding axiomatically and empirically
(\S\ref{sec:background}). Both routes are \emph{intrinsic} by nature: nothing
external says what a good similarity score would be, so a measure can only be
held against other measures, against its own stability, or against perturbations
whose importance is argued rather than observed. That is why the choice among
these measures is still open after a decade.

Answering these questions calls for a change of method rather than a refinement
of it: an \emph{extrinsic} evaluation, one that holds a measure against an
account of how the conditions ought to be arranged that owes nothing to any
system being compared. Such accounts are rare, which is why the literature has
done without one. Showing that a choice \emph{changes} a conclusion requires no
more than varying it; showing that a conclusion is \emph{wrong} requires knowing
what it should have been, and in most domains the structure the conditions ought
to exhibit is exactly what is being estimated.

Speech supplies a domain where it is not. The acoustic arrangement of vowel
categories is a genuine low-dimensional geometry, recoverable from the waveform
by signal processing that shares nothing with the models under evaluation,
documented for over seventy years \citep{petersonbarney1952}, and different in
every language. \textbf{This paper uses it to measure how much of a comparative
conclusion is fixed by the condition set rather than by the data}, in a setting
where a choice that changes the answer can be shown to make it wrong about the
claim as stated, not merely different. \S\ref{sec:instrument} describes the
instrument and its scale, on which a similarity score of $0.87$ can be called large or small
rather than merely positive.

Every one of the evaluations above varies the measure and holds the set of
conditions fixed, which is less an oversight than a consequence of the framing:
the invariances of a measure are properties of what it does
\emph{once the $N \times N$ matrix has been formed}, so no invariance argument
reaches a decision taken before that matrix exists. Both decisions that fix the
condition set are taken before it. Each is mandatory, unconventioned, almost
never reported, and, as far as we are aware, unmeasured. That analytic choices
of this kind go undeclared, and change conclusions when they do, is documented
at scale in neighbouring fields: almost every one of 241 surveyed fMRI studies
used a unique analysis pipeline \citep{carp2012}, and seventy teams given one
dataset and the same nine hypotheses chose seventy different workflows and
reported different results \citep{botvinik2020}. A filter can even attenuate the
effect studied while creating a spurious one of opposite sign
\citep{tanner2015}.

Our own question makes both decisions visible. Asking whether a self-supervised
speech model recovers that geometry better than a non-learned spectral
representation does requires first deciding which stretches of audio count as
instances of a vowel, and then which vowel categories those instances are
grouped into. Neither is a matter of presentation: how many degrees of vowel
height to recognise, or whether a diphthong is one condition or two, are
contested theoretical questions on which specialists disagree
(Appendix~\ref{app:inventories}). Each choice is therefore constrained but not
determined, which is the dangerous case: one recognised as arbitrary would be
reported, whereas one that is constrained but underdetermined is mistaken for a
fact about the data.

It turns out to dominate. Changing which vowel categories are compared, at a
fixed number of them, moves the gap between two representations by
\CompSepTypical\ on average: more than two and a half times the effect being
measured, \CompEffectKFive. Changing which stretches of audio count as instances
moves the gap in one language of fifteen from \ThrPlBefore\ to \ThrPlAfter. Changing
\emph{how many} categories are compared, by contrast, has no measurable effect
at all: from five to twenty-eight the gap moves by \CompRangeKRounded, of the
order of its own standard error. What matters is neither how many conditions
there are nor how widely they are spread through the reference space, but which
ones they are. \S\ref{sec:results} gives the measurements.

\paragraph{Contributions.}
We treat the condition set as an object of measurement rather than as a given.
Against a reference owing nothing to the systems compared, we quantify what each
decision determines and which aspect of the set is responsible, show the effect
to be structured rather than noisy, establish that a damaging selection cannot be
recognised without recomputation, and report which claims remain identified once
both decisions vary. Each answer is about correctness and not only stability. We
also report what our pre-registered protocol caught
(Appendix~\ref{app:caught}).

\section{Background: measuring representational similarity}
\label{sec:background}

The measures that answer the problem of \S\ref{sec:intro} compare the internal
structure of each system rather than the two sets of vectors: how the things
being represented are arranged relative to one another, within one system and
then within the other. They form one family among several \citep{klabunde2025},
and the one on which the methodological debate has concentrated. This section
fixes notation for the three we use, and locates within them the decision this
paper measures. Approaches that instead fit a map between the two systems, or
compare them by their behaviour, are discussed in
Appendix~\ref{app:alternatives}.

The rows of that arrangement are the \emph{conditions} of \S\ref{sec:intro}, and
what a row is varies by tradition. In much of the machine-learning literature a
row is a single input (one image, one utterance) and the only decision is which
inputs enter \citep{kornblith2019,raghu2017}. In the neuroscience tradition
from which these measures come, a row is an experimental condition summarised
over repeated trials \citep{kriegeskorte2008,nili2014}, and in analyses at the
level of categories (object classes, word types, phonetic categories) a
row pools every input carrying the same label. Where rows aggregate, both
decisions apply: which inputs are admitted, and how the admitted inputs are
grouped. Where a row is a single input, only the first does.

Nor is the grouping always the same on both sides. In model-to-brain
comparisons the neural side is averaged over trials while the model side is
deterministic, so the two sides are pooled by different rules as a matter of
course, so that the estimator on one side is not the estimator on the other.

Write $\mathbf{X} \in \mathbb{R}^{N \times p}$ and
$\mathbf{Y} \in \mathbb{R}^{N \times q}$ for the two systems at the level of
conditions:
$N$ conditions, one per row, in $p$ and $q$ dimensions respectively. The
comparison is now between two objects of size $N \times N$, one per system, and
$N$ is the size of the condition set rather than the number of inputs.

RSA \citep{kriegeskorte2008} builds, for each system, the $N \times N$
representational dissimilarity matrix (RDM) of distances between its conditions,
and correlates the two, typically with Spearman's $\rho_s$, so that no linear
relation between the two distance scales need be assumed:
\begin{equation}
\left(\mathrm{RDM}_{\mathbf{X}}\right)_{ij} = d(\mathbf{x}_i, \mathbf{x}_j),
\qquad
\mathrm{RSA}(\mathbf{X},\mathbf{Y}) =
\rho_s\!\left(\mathrm{vec}^{\triangle}\mathrm{RDM}_{\mathbf{X}},\;
              \mathrm{vec}^{\triangle}\mathrm{RDM}_{\mathbf{Y}}\right),
\label{eq:rsa}
\end{equation}
where $\mathbf{x}_i$ is row $i$ of $\mathbf{X}$ and $\mathrm{vec}^{\triangle}$
vectorises the strict upper triangle of a symmetric matrix, $m = N(N-1)/2$
entries, one per pair of conditions; $m$ bounds what the correlation can
resolve.

Linear CKA \citep{kornblith2019} builds a different $N \times N$ object from the
same rows, the Gram matrix of inner products between conditions, and normalises
the alignment of the two through HSIC, which measures that alignment once the
mean has been removed from each. Orthogonal Procrustes \citep{williams2021}
takes a third form, minimising
$\lVert \mathbf{X}\mathbf{Q} - \mathbf{Y} \rVert_F$ over rotations
$\mathbf{Q}$; since $p \neq q$ in general, the two must first be brought to a
common dimension, at a projection rank the definition does not supply
(Appendix~\ref{app:measuredefs}). The three also differ in what they leave to
the analyst: RSA fixes neither the distance $d$ nor the statistic comparing the
triangles, which CKA and Procrustes fix by definition.

Every object in these definitions is indexed by conditions and not by
dimensions, which is what makes each invariant to permutation of dimensions by
construction, and to orthogonal rotation whenever the underlying proximity is
Euclidean. That invariance is bought at a price, and the price is the subject of
this paper: the coordinate space has been discarded and only the arrangement
among conditions retained, so everything downstream depends on \emph{how that
arrangement was built}, and the condition set is its first ingredient. Which
inputs enter, and how they are grouped, is not a detail of the procedure: it is
the object the procedure operates on.

The methodological debate around these measures is, with no exception we are
aware of, a debate about invariance, and for the reason \S\ref{sec:intro} gives:
with no external criterion for what a good similarity score would be, a measure
can only be held against controls internal to the comparison. RSA was introduced
to compare measurement modalities with no common units at all
\citep{kriegeskorte2008} and supplied with a reliability analysis shortly after
\citep{nili2014}. CCA and its variants sought invariance to any invertible
linear map \citep{raghu2017,morcos2018}; \citet{kornblith2019} showed this to be
too permissive, since discarding the distinction between high- and low-variance
directions makes trained and untrained networks resemble one another, and
proposed linear CKA with the invariance weakened to the orthogonal and isotropic
cases. \citet{williams2021} responded to both, on the grounds that neither RSA
nor CKA is a metric and so neither licenses clustering models or averaging over
them. Two later contributions replaced axioms by experiment while keeping the
controls internal: \citet{ding2021} perturbed representations in ways known on
independent grounds to change what a network computes and found that the
measures differ in whether they register the change at all, so that the answer
to a comparative question can depend on which measure was used, and
\citet{davari2023} showed that CKA can be driven to arbitrary values without
materially changing what a network computes.

Every one of those proposals concerns what a measure does to the $N \times N$
matrix. What none of them leaves open is the condition set, which each receives
already formed.

\section{An instrument with a known answer}
\label{sec:instrument}

To measure a choice no invariance argument can reach, we need a case in which
the arrangement the conditions ought to have is known before any model is
consulted. Such cases are uncommon: in most domains where these measures are
applied, the structure the conditions should exhibit is what is being estimated.
Vowels offer one. Vowel categories occupy distinct and stable regions of a
two-dimensional acoustic plane, the vowel triangle, documented since the 1950s
\citep{petersonbarney1952} and used in phonetics as a measurement tool rather
than a theoretical idealisation. Its coordinates are the first two
\emph{formants}, $F_1$ and $F_2$, the two lowest resonances of the vocal tract,
which track tongue height and backness and are recovered from the waveform by
standard signal processing.

What this yields is a \emph{geometry} and not a set of labels. Positions in the
$(F_1, F_2)$ plane admit distances, so for any set of vowel categories a
pairwise structure can be written down against which a model's own may be
compared, which class labels or a proxy task could not supply. It comes from the
same recordings the models receive, by a route sharing no component with any of
them: the alignment locating the tokens is produced by a trained model
\citep{mcauliffe2017}, but one not among the systems evaluated, so its errors are
common to all rather than favouring any.

We call this an \emph{external reference} and not a ground truth because it is
itself a measurement, with its own assumptions and errors; what makes it usable
is not that it is beyond doubt but that it is fixed before any model is
consulted. It does not state what a speech representation ought to encode, only
one measurable property of the signal against which a comparative claim can be
evaluated.

Because that arrangement is known, a choice that changes the answer can be shown
to make it wrong about the claim as stated, and not merely different; and because inventories differ, the question of
\S\ref{sec:intro} can be asked once per language.

\paragraph{Data and protocol.}
We use fifteen of the 102 languages of FLEURS \citep{conneau2023}, a read-speech
corpus with content parallel across languages, chosen to span inventories of
five to twenty-eight categories and to vary in tone and in family
(Appendix~\ref{app:languages}); the corpus carries no speaker labels, which
constrains what can be controlled for.

Forced alignment \citep{mcauliffe2017} gives a phone label and a time interval
for every vowel token, formants are estimated with Praat \citep{boersma2024},
and two levels of aggregation reduce each system to one vector per category:
frames are mean-pooled within a token, then tokens sharing a label are
summarised by their mean on the representation side and their coordinate-wise
median on the reference side. The two sides therefore use different estimators,
each with a local justification and the pair with none, which
\S\ref{sec:aggregation} measures. Comparing a representation with the reference
is then RSA as defined in \eqref{eq:rsa}, its two open choices fixed to
cosine distance on the representation side, Euclidean on the reference side, and
Spearman's $\rho$ between the vectorised triangles: the default throughout, from
which each subsection of \S\ref{sec:results} varies one step
(Appendix~\ref{app:protocol}).

\paragraph{Representations compared.}
Twenty-seven representations are evaluated, chosen so that the scale defined
below has known points along its whole length rather than only at the top.
Three families are the objects of the comparison: six self-supervised models,
one of them pre-trained on music, two encoders trained or fine-tuned on
transcriptions, and four neural audio codecs. Two fix the scale from below: four
non-learned representations, and nine randomly initialised controls
matching the trained architectures layer for layer, which isolate what the
architecture contributes before any training. The last is a floor: one model
trained on images, run at two frame rates, which shares the architecture family
but not the modality. Appendix~\ref{app:models} gives each checkpoint with its
revision hash, the architecture and initialisation of every control, and how
frames are obtained from the vision model.

\paragraph{An anchored scale.}
A similarity score has no natural units, which is tolerable when the only
question is which of two models scores higher. Here it is not: we must say
whether the effect of a pipeline choice is large \emph{relative to the claimed
effect}, and without a common yardstick the sentence \emph{the choice of
categories moves the result more than pre-training does} cannot even be
evaluated.

The reference provides that yardstick. Table~\ref{tab:scale} reports, per family
of representations, the median over the fifteen languages of the RSA score in
the default configuration just described. Three rows are controls rather than
models, and they are what makes the others readable: the reference passed
through the representation side of the pipeline and compared with itself, which
scores \ScaleCeiling\ and which no representation can better; the trained
architectures with their weights left at initialisation; and the model trained
on images, which shares the architecture family but has never heard speech.
Together they bound the axis from above, from the middle and from below.

\begin{table}[t]
\centering
\small
\caption{The anchored scale: median $\rho$ over the fifteen languages, one row per representation family.}
\label{tab:scale}
\begin{tabular}{lll}
\toprule
$\rho$ & anchor & content \\
\midrule
\ScaleCeiling & reference ceiling & identity control: the geometry against itself \\
0.9003 & pre-trained speech and audio models & 8 models: 6 self-supervised, 2 supervised \\
0.8535 & non-learned spectral features & 3 baselines \\
0.8421 & learned compression & 4 neural codecs \\
0.7571 & raw formant measurements & $F_1$--$F_4$ as a feature vector \\
\ScaleRandom & architecture alone & 9 randomly initialised controls \\
0.3298 & out of modality & 1 vision checkpoint at 2 frame rates \\
\bottomrule
\end{tabular}
\end{table}

We express every effect as a percentage of the interval between the first
control and the second: the distance between an untrained architecture and exact
recovery of the geometry. Both ends are controls and not models, deliberately. A
unit defined as the distance between two named models would be the very quantity
the scale is used to judge, and every effect would be reported as a fraction of
itself. That interval is the unit of \S\ref{sec:results} throughout, the
instrument that makes one pipeline choice comparable with another and with the
effect being claimed.

Anchoring that scale on RSA is itself a choice, made for resolution: each
measure has an interval of its own, RSA's is the widest of the three at
\MetIntervalRsa{} against \MetIntervalCka{} and \MetIntervalProc{}
(Table~\ref{tab:metric_agreement}), and the widest interval is the one on which
two effects of similar size are least likely to be confused. The choice is not
neutral, and \S\ref{sec:measures} measures what it is worth: read each on its
own interval, the three return the pre-training gap with the same sign and order
of magnitude.

\paragraph{How much of the axis is noise.}
The reference is estimated, so its reproducibility bounds what any score against
it can mean. A split-half reliability of the reference matrix, by utterance with
the Spearman--Brown correction \citep{nili2014}, runs from \CeilMin\ to
\CeilMax\ across languages with a median of \CeilMedian, and no
model--language cell exceeds the ceiling of its language: what
\S\ref{sec:results} reports are differences and not noise. One qualification
limits what we recommend. Reliability is not validity: a systematic error, an
octave jump made consistently by the formant tracker, reproduces in both halves
and leaves the ceiling untouched, so a high ceiling certifies that the reference
is stable, not that it is right, and the control of
Appendix~\ref{app:checklist} inherits that blind spot.

\paragraph{Bifurcations.}
Every step just described involves a choice the procedure requires and nothing
determines: which tokens are admitted, which labels kept apart, which estimator
summarises a category, which distances are taken. We call such a choice a
\emph{bifurcation} when it is mandatory, unconventioned, rarely reported and,
decisively, \emph{asymmetric}: when it does not affect the compared systems
alike.

That last property carries the argument. A choice shifting every score equally is
a nuisance parameter, changing values and leaving conclusions alone; one that
affects the compared systems differently changes rankings. Only the second
threatens identification, so we report for every bifurcation a
\emph{common effect}, the mean shift across systems, and a \emph{differential
effect}, the variance of that shift between them. Only the second bears on what
may be concluded.

%
%
%
\section{Results}
\label{sec:results}

\begin{figure}[t]
\centering
\includegraphics[width=\linewidth]{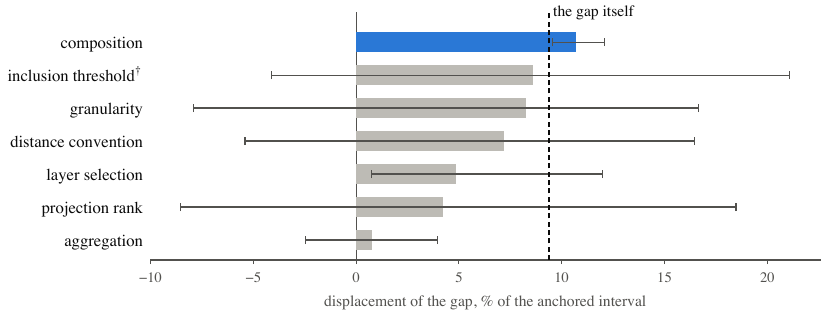}
\caption{Each undeclared choice, and the effect it is meant to help measure.
Bars give the displacement of the pre-training gap in percent of the anchored
interval (\S\ref{sec:instrument}); the vertical line is the gap itself. The
dagger marks the sweep run over eight representations, not twenty-seven.}
\label{fig:bifurcations}
\end{figure}

We begin with the choice the literature has debated, then turn to the ones it
has not. The unit of replication differs by bifurcation: choices applying to
every language are bootstrapped over the fifteen, whereas composition is
evaluated over repeated draws within one language, so the two intervals are not
comparable estimates of the same uncertainty and we do not read them as such. Figure~\ref{fig:bifurcations} is the shape of what follows: those choices
displace the gap between pre-trained models and non-learned spectral features,
the largest by more than the gap itself.

\subsection{What the measures decide, and what they leave open}
\label{sec:measures}

The literature's question is which of these measures to use, so we ask first
what that choice is worth. RSA, linear CKA and Procrustes at matched rank are
applied to the same representations, conditions and reference, and compared on
the orderings they produce.

\begin{table}[t]
\centering
\small
\caption{The three measures on one question. Top: rank correlation between two measures of the ranking of representations. Bottom: the gap between pre-trained speech and audio models and non-learned spectral features under each measure, on that measure's own anchored interval.}
\label{tab:metric_agreement}
\begin{tabular}{lccc}
\toprule
pair of measures & 27 representations & 16 trained & 16 trained, English only \\
\midrule
RSA / CKA & \MetAgreeRsaCkaAll\ {\tiny \MetAgreeRsaCkaAllCI} & \MetAgreeRsaCkaTrained\ {\tiny \MetAgreeRsaCkaTrainedCI} & \MetAgreeRsaCkaEnTrained \\
RSA / Procrustes & \MetAgreeRsaProcAll\ {\tiny \MetAgreeRsaProcAllCI} & \MetAgreeRsaProcTrained\ {\tiny \MetAgreeRsaProcTrainedCI} & \MetAgreeRsaProcEnTrained \\
CKA / Procrustes & \MetAgreeCkaProcAll\ {\tiny \MetAgreeCkaProcAllCI} & \MetAgreeCkaProcTrained\ {\tiny \MetAgreeCkaProcTrainedCI} & \MetAgreeCkaProcEnTrained \\
\midrule \textit{measure} & \textit{floor -- ceiling} & \textit{gap} & \textit{\% of its own interval} \\
RSA & \MetFloorRsa--\MetCeilRsa & \MetGapRsa & \makecell[t]{\MetGapRsaPct \\[-3pt] {\tiny \MetGapRsaPctCI}\rule[-2pt]{0pt}{2pt}} \\
linear CKA & \MetFloorCka--\MetCeilCka & \MetGapCka & \makecell[t]{\MetGapCkaPct \\[-3pt] {\tiny \MetGapCkaPctCI}\rule[-2pt]{0pt}{2pt}} \\
Procrustes, rank 2 & \MetFloorProc--\MetCeilProc & \MetGapProc & \makecell[t]{\MetGapProcPct \\[-3pt] {\tiny \MetGapProcPctCI}\rule[-2pt]{0pt}{2pt}} \\
\bottomrule
\end{tabular}
\end{table}

Less than the literature would lead one to expect. That these measures can
answer the same question differently is well established
\citep{kornblith2019,ding2021,davari2023}; here it does not reach the ordering.
The three rank the twenty-seven representations alike, and the gap between
pre-trained models and non-learned spectral features keeps its sign and its
order of magnitude under all three, each read on its own anchored interval
(Table~\ref{tab:metric_agreement}). They part company where comparisons are
actually made: among the sixteen trained and spectral representations, without
the controls, agreement falls markedly. Much of it comes from the distance
between trained models and controls rather than from the ordering among trained
models, and CKA moves those most, promoting a codec and a music-pretrained model
and demoting a supervised encoder. None of this makes the three
interchangeable: they agree on the coarse ordering here and disagree
substantially once the controls are removed.

That reassurance is conditional, and on the kind of choice this paper is about.
A measure is not a procedure: RSA fixes neither the distance nor the statistic
comparing the triangles, and Procrustes at matched rank requires a projection
rank (\S\ref{sec:background}). Varying those, four of five rank-by-domain
configurations change verdict between the two distance conventions, in both
directions, and the two factors interact (Table~\ref{tab:rank-verdicts},
Appendix~\ref{app:tables}). The reversal is not an artefact of our own pipeline:
under the same change, a quantity that the convention cannot reach does not move
at all (Appendix~\ref{app:tables}). The sensitivity this paper is about is
therefore already visible in the question the literature has been asking. The
measures agree, until one of the choices they leave open is made differently.

One limit no measure escapes. A rank correlation over $k$ conditions moves on a
grid of spacing $12/\bigl(m(m^2-1)\bigr)$ with $m = k(k-1)/2$, so below some $k$
the difference reported is smaller than the step between attainable values: at
five categories every observed value falls on a grid point and \MakeLowercase{\QuantTied}
models share one. The criterion is operational, not a rejection of small
inventories, and \S\ref{sec:granularity} gives the case that fails it.

\subsection{How the inputs are grouped}
\label{sec:composition}

An analyst comparing two systems over $k$ vowel categories must choose which
$k$; nothing in the procedure fixes the choice and it is rarely reported. Drawing
five hundred subsets at each of six cardinalities separates the effect of the
size of the set from that of its composition (Table~\ref{tab:sigma-by-k},
Appendix~\ref{app:tables}). Size has no measurable effect: from five categories
to twenty-eight the gap moves by less than its own standard error. Composition
has a large one: two analysts drawing five categories each report gaps separated
by \CompSepTypical\ on average, against an effect of \CompEffectKFive\ that
both are trying to measure. We report the two quantities separately rather than
their ratio, whose denominator is not distinguishable from zero and whose value
depends on two undeclared choices (Appendix~\ref{app:tables}).

Random subsets are not what an analyst would pick, so we restrict the draws to
the dispersion of the fifteen attested inventories: that removes a fifth of it at
five categories and leaves the rest twice the effect, so the result is not
an artefact of implausible sets. Nor would a summary serve in its place: four
coverage statistics account for \CovRsqPooled\ of the pooled variance and
cardinality alone for none of it (Appendix~\ref{app:composition}), so the set
has to be reported category by category.

What matters for published work is where an inventory somebody built falls among
those they could have built instead. The twelve-category English merge table used
here was built for this project
(Appendix~\ref{app:composition}); it falls at percentile \PctEnSubsets\ among
five hundred random subsets of its size and at \PctEnMerges\ among merges
matched for the profile of category sizes, while four languages whose inventory
changed without a hand-built table fall between \PctOthersLow\ and
\PctOthersHigh, the lower bound of the English interval lying above all four. A
hand-built inventory is therefore not one arrangement among many: it lies where
the distribution is thin, which is what makes it unrepresentative of the choices
a reader might have made.

\label{sec:granularity}
Which categories are compared is half of the grouping decision; how finely they
are cut is the other, and merging is the commonest manipulation of an inventory,
always on the ground that finer distinctions are unreliable. Merging that raised
every score equally would be a nuisance parameter. Merging the twenty-eight
English categories into twelve raises every score, by \GranCommon\ on average,
and does not raise them equally. The shift is monotone in the quality of the representation
($\rho = \GranRho$, $p = \GranRhoP$): the six weakest representations gain
\GranWeakSix, the six strongest \GranStrongSix, and the spectral baselines lose
ground. A coarse inventory flatters weak models. The consequence is that the gap
between a strong self-supervised model and its randomly initialised
counterparts falls from \GranHubertRandomFine\ to \GranHubertRandomCoarse: the
sign of the comparison is preserved, and the margin supporting it depends on a
convention that is rarely declared.

\subsection{Which inputs are admitted}
\label{sec:threshold}

Tokens shorter than some duration are routinely discarded, on the ground that a
formant estimate over too few periods is unreliable. The threshold is never the
same twice and never reported. We sweep it from \ThrMin\ to \ThrMax\
milliseconds, extracting once at zero and filtering afterwards so that the sets
are nested.

The rule changes what is concluded. The ranking of languages survives only
partly, and in Polish, the only language in which
the spectral baseline is ahead, the threshold moves the gap from
\ThrPlBefore\ to \ThrPlAfter. About half of what the threshold does is to move the compared systems
relative to one another rather than together.

A filter's bias is measurable without recomputing anything: one compares the
distribution of categories among the tokens kept and those discarded. We
pre-registered the prediction that this quantity would predict the threshold's
differential effect, and that both would be largest in the language included in
the corpus for its extreme vowel reduction.

The prediction fails twice. Russian is indeed the most heavily cut language and
first on both measures of bias, yet its differential effect ranks
\DiagRuRank\ of fifteen, and across the fifteen languages the correlation
between measured bias and differential effect is \DiagCorr\ ($p = \DiagCorrP$).
The Russian case keeps its force as a matter of construct validity, since the
filter removes precisely the phenomenon the language was included for. But the
one language in which the damage was visible in advance is not the one in which
it was worst, and the cheap diagnostic does not substitute for the sweep.

\subsection{What the analyst does downstream}
\label{sec:aggregation}

Once the condition set is fixed, two choices remain: how a condition's tokens are
summarised and which layer of a deep model is reported. Neither belongs to the
set and neither is usually declared. The two estimators of
\S\ref{sec:instrument}, a mean on the representation side and a coordinate-wise
median on the reference side, are a bifurcation of the kind this paper is about,
found in our own work; we recompute the representation side with the reference's
estimator (Table~\ref{tab:aggregation}, Appendix~\ref{app:tables}).

Matching the two estimators displaces the comparison by \AggDifferentialPct\ of
the anchored interval and lowers every anchor. We had predicted that the gap
would narrow, since the spectral baselines carry the heavier-tailed
within-category distributions and should gain most from a robust centroid. The
premise is correct and the gap nonetheless widens: the reference is itself a
coordinate-wise median over formant measurements, and the tokens it discards are
not those lying far from a centroid in a spectral space, so matching the name of
an estimator does not match its effect. Neither configuration is principled and
we report both. As with the inclusion threshold, a quantity available without
recomputation orders the systems correctly and fails to predict the consequence.

\label{sec:layer}
Where a model has many layers and a baseline one, reporting each at its best
layer compares a maximum over many draws with a maximum over one, an asymmetric
multiple-comparison advantage that grows with the number of layers and so cancels
in no comparison. Matched relative depth is one control among several; we use it
to size that advantage, not to claim it removed.

Of the \LayerPairs\ paired comparisons, \LayerSurviving\ survive the change
of protocol. The inflation is \LayerInflationPct\ of the anchored interval against a
median gap of \LayerGap, so selection accounts for \LayerShare\ of the reported
margin. This is the one bifurcation we can test outside speech, and there the
reference is exact rather than measured: on dSprites, whose images are generated
from known factors, a baseline contests \VisContestedMatched\ of \VisPairs\
comparisons between three trained vision models and four unlearned ones at
matched relative depth, where raw pixels are ahead of all three
(Appendix~\ref{app:vision}).

\subsection{What survives}
\label{sec:hierarchy}

Each effect above was reported in the quantity natural to its analysis, and those
quantities are not commensurable, so we ask one question of each: how far does it
move the gap between pre-trained models and non-learned spectral features.
Figure~\ref{fig:bifurcations} gave the shape, Table~\ref{tab:closing} the values
(Appendix~\ref{app:tables}).

One choice separates and five are comparable. Only composition has an interval
excluding the others; the inclusion threshold, granularity, the distance
convention, layer selection and projection rank follow, ordered by estimate, with overlapping
intervals all containing zero but the layer. Aggregation is negligible. They
support reading \emph{composition first, alone in separating; then a group of
five of comparable magnitude, the choice of measure among them; then
aggregation}, not a strict ordering of seven lines.

Two qualifications belong with that comparison, beyond the unit of replication
noted above, which also leaves composition without a multilingual interval since
only one language has twenty-eight categories. Two lines change rank once brought
to this unit, aggregation falling and granularity rising, because a spread
between models and a displacement of the gap are two quantities
(Appendix~\ref{app:displacement}). The dominance is not an artefact of the
protocol: standardising the reference once over the full inventory rather than on
each subset gives \CloseCompIqrZfullPct\ instead of \CloseCompIqrPct, with the
curve unchanged in shape.

That comparison measures displacement, whereas a published comparison usually
reports an ordering, which may be more robust. We compute the rank correlation of
the twenty-seven representations between the extremes of each choice and between
pairs of draws.

\begin{table}[t]
\centering
\small
\caption{What survives each choice: rank correlation of a ranking between the two extremes of the choice, with a bootstrap interval and the number of items it is computed on.}
\label{tab:ranking_survival}
\resizebox{\textwidth}{!}{%
\begin{tabular}{lcccc}
\toprule
choice & models & trained models & languages, by level & languages, by gap \\
\midrule
inclusion threshold, 0 vs 100 ms & \HierModelsThr\ {\tiny \HierModelsThrCI}, $n = \HierModelsThrN$ & --- & +0.786 {\tiny [+0.41, +0.95]}, $n = 15$ & +0.507 {\tiny [-0.07, +0.88]}, $n = 15$ \\
distance convention & \HierModelsMet\ {\tiny \HierModelsMetCI}, $n = \HierModelsMetN$ & --- & +0.432 {\tiny [-0.14, +0.81]}, $n = 15$ & +0.518 {\tiny [+0.04, +0.86]}, $n = 15$ \\
\midrule \multicolumn{5}{l}{\textit{English only, like for like}} \\
composition, $k = 20$ & \RankDrawsKTwenty\ {\tiny \RankDrawsKTwentyCI}, $n = 27$ & \RankDrawsKTwentyTrained\ {\tiny \RankDrawsKTwentyTrainedCI}, $n = 16$ & --- & --- \\
composition, $k = 14$ & \RankDrawsKFourteen\ {\tiny \RankDrawsKFourteenCI}, $n = 27$ & \RankDrawsKFourteenTrained\ {\tiny \RankDrawsKFourteenTrainedCI}, $n = 16$ & --- & --- \\
measure, RSA vs Procrustes & \MetAgreeRsaProcEnAll, $n = 27$ & \MetAgreeRsaProcEnTrained, $n = 16$ & --- & --- \\
composition, $k = 5$ & \RankDrawsKFive\ {\tiny \RankDrawsKFiveCI}, $n = 27$ & \RankDrawsKFiveTrained\ {\tiny \RankDrawsKFiveTrainedCI}, $n = 16$ & --- & --- \\
\bottomrule
\end{tabular}
}
\end{table}

Orderings survive more than values do, but not uniformly. The ranking of models
correlates above \HierModelsThr\ across both undeclared choices and at
\RankDrawsKFourteen\ between two draws of fourteen categories. Restricted to the
sixteen trained and spectral representations the last falls to \RankDrawsKFourteenTrained,
which is still higher than two measures agree with each other on the same
language. At five categories the ordering collapses: \RankDrawsKFiveTrained\
between draws over the trained representations, with a quarter of draw pairs
below \RankDrawsKFiveQuartile. That regime is not hypothetical: Spanish,
Mandarin and Polish have inventories of that size.

The thesis is therefore best stated by regime. On inventories of typical
size, the composition of the condition set moves the values more than the choice
of measure, and disturbs the ordering less. Only on small inventories does it
disturb the ordering more than the choice of measure.

In this setting comparisons between models are identified and comparisons across
languages are not, since two unrelated undeclared choices each suffice to
scramble that ranking.

\section{Conclusion}
\label{sec:conclusion}

A decade of work on representational similarity has gone into choosing among
measures; that is not where the leverage is. The three we compared rank
twenty-seven representations alike and agree on the pre-training gap more closely
than the choices we measured differ, while the composition of the condition set
moves that gap by more than the gap itself. What is compared beats how it is
compared.

The argument is not against the invariance literature, which answered its own
question, but about ordering. A measure's invariances are properties of what it
does once the $N \times N$ matrix exists, so no result there reaches a decision
taken before that matrix is formed, as both decisions fixing the condition set
are. A measure can be invariant to everything it was designed
to be invariant to yet return whatever the analyst's inventory determines.

Three things follow for practice. A comparative claim should report its condition
set category by category, since no summary of the set stands in for it; report
the rule that admitted its inputs, and sweep it, since no cheap diagnostic
predicts what that rule will do to a comparison; and compare systems
at matched depth, or say that it did not. Appendix~\ref{app:checklist} states
these alongside our own record, offered as a base rate and not as evidence of
rigour: pre-registration, a bug journal and a test suite still left the defects
listed in Appendix~\ref{app:caught}.

\paragraph{What this does not establish.}
The two decisions differ in reach: grouping bears on work in which a row
aggregates several inputs, admission wherever these measures are used at all. Two
of our results rest on English, the only language of our sample in which the
space of condition sets can be explored widely and whose hand-built inventory is
itself unusual among those we sample (\S\ref{sec:composition}); only the
layer-selection artefact was tested in a second modality, and at least two
further bifurcations meet our criterion unexamined. The magnitudes are therefore
established for vowel geometry in multilingual speech; the structural claim above
is not so confined (Appendix~\ref{app:limitations}).

What made any of this measurable was a criterion external to the comparison.
Intrinsic validation can establish that two measures disagree, not which is
wrong, nor see a choice taken before any measure is applied. Domains where the
arrangement of the conditions is known independently of the systems compared are
rare, but not confined to speech: the layer-selection artefact
reproduces, and is larger, where generative factors supply the reference exactly,
so that the objection to our own reference cannot arise
(Appendix~\ref{app:vision}). We would rather the field looked for more such
domains than a better measure.

\section*{Reproducibility Statement}

Every measured value in this paper, in the text, tables and captions alike, is
generated programmatically from the stored result files rather than transcribed
by hand, so the prose cannot drift from the results it reports; the few
exceptions are fixed protocol constants and a handful of figures from
exploratory runs whose scripts write no result file. Each value is bound to an
artefact and its checksum, and the generation chain refuses to build when two
names are bound to the same value of the same artefact, when a table declares a
quantity another table declares from a different source, or when a percentage
cannot be recomputed from its operands. The pipeline was frozen after the
last defect recorded in Appendix~\ref{app:caught} and every value regenerated
from it, so no number reported here predates a correction; the largest of those
corrections moved the anchored interval in its fifth decimal and changed no
conclusion. The extraction, analysis and figure code, together with the scripts
that regenerate every reported value, are included in the
\ifdefined\deanonymized\else anonymous \fi supplementary material.

Appendix~\ref{app:preregistration} states which of our pre-registered
predictions can be checked from outside the project and which cannot, and
Appendix~\ref{app:caught} gives the defect record in full. The protocol is in
Appendix~\ref{app:protocol}, the language sample in
Appendix~\ref{app:languages}, and the checkpoints with their revision hashes in
Appendix~\ref{app:models}. One sweep is narrower than the others for a reason
that is recoverable: the stored representations predate the change that records
segment durations, so extending the duration threshold from eight
representations to twenty-seven requires re-extracting nineteen models over
fifteen languages, of the order of three GPU-days. That is a decision about
time, not a technical obstacle, and we say so.

\section*{Ethics Statement}

This work analyses public corpora and public model checkpoints and collects no
new data from human participants. FLEURS is used as published, with no
annotation added; it carries no speaker identifiers, which bounds what could be
inferred about any individual from our artefacts.

One of our results bears on a class of claim with equity implications.
Comparisons of how well a representation serves one language against another are
precisely the comparisons \S\ref{sec:hierarchy} finds not to be identified under
choices that go undeclared. Statements that a multilingual model represents some
languages as well as others, or serves a low-resource language adequately, are
therefore more fragile than their published form suggests, and we would not want
the instrument of \S\ref{sec:instrument} read as licensing them. The remedy is
the one this paper recommends throughout: report the condition set, and the rule
that admitted the inputs.

\section*{AI Use Statement}

The authors used a large language model as a writing aid, for drafting and
revising prose and for checking internal consistency across sections, and as a
coding assistant in the preparation of the analysis code. It was not used to
generate the research ideas or the hypotheses tested here, and no number in the
paper is produced by a language model: every value is computed by the pipeline
described in Appendix~\ref{app:protocol} from stored artefacts, and bound to
them as the Reproducibility Statement describes. The authors take full
responsibility for the content, including every claim and citation.


\bibliographystyle{iclr2027_conference}
\bibliography{metric_comparison_evaluation}

\begin{thebibliography}{24}
\providecommand{\natexlab}[1]{#1}
\providecommand{\url}[1]{\texttt{#1}}
\expandafter\ifx\csname urlstyle\endcsname\relax
  \providecommand{\doi}[1]{doi: #1}\else
  \providecommand{\doi}{doi: \begingroup \urlstyle{rm}\Url}\fi

\bibitem[Alain \& Bengio(2016)Alain and Bengio]{alain2016}
Guillaume Alain and Yoshua Bengio.
\newblock Understanding intermediate layers using linear classifier probes.
\newblock \emph{arXiv preprint arXiv:1610.01644}, 2016.

\bibitem[Boersma \& Weenink(2021)Boersma and Weenink]{boersma2024}
Paul Boersma and David Weenink.
\newblock Praat: Doing phonetics by computer, 2021.
\newblock Computer program, version 6.1.38.

\bibitem[Botvinik-Nezer et~al.(2020)Botvinik-Nezer, Holzmeister, Camerer,
  Dreber, Huber, Johannesson, Kirchler, Iwanir, Mumford, Adcock,
  et~al.]{botvinik2020}
Rotem Botvinik-Nezer, Felix Holzmeister, Colin~F. Camerer, Anna Dreber, Juergen
  Huber, Magnus Johannesson, Michael Kirchler, Roni Iwanir, Jeanette~A.
  Mumford, R.~Alison Adcock, et~al.
\newblock Variability in the analysis of a single neuroimaging dataset by many
  teams.
\newblock \emph{Nature}, 582\penalty0 (7810):\penalty0 84--88, 2020.
\newblock \doi{10.1038/s41586-020-2314-9}.

\bibitem[Carp(2012)]{carp2012}
Joshua Carp.
\newblock The secret lives of experiments: Methods reporting in the {fMRI}
  literature.
\newblock \emph{NeuroImage}, 63\penalty0 (1):\penalty0 289--300, 2012.
\newblock \doi{10.1016/j.neuroimage.2012.07.004}.

\bibitem[Conneau et~al.(2023)Conneau, Ma, Khanuja, Zhang, Axelrod, Dalmia,
  Riesa, Rivera, and Bapna]{conneau2023}
Alexis Conneau, Min Ma, Simran Khanuja, Yu~Zhang, Vera Axelrod, Siddharth
  Dalmia, Jason Riesa, Clara Rivera, and Ankur Bapna.
\newblock {FLEURS}: Few-shot learning evaluation of universal representations
  of speech.
\newblock In \emph{2022 IEEE Spoken Language Technology Workshop (SLT)}, pp.\
  798--805. IEEE, 2023.
\newblock \doi{10.1109/SLT54892.2023.10023141}.

\bibitem[Davari et~al.(2023)Davari, Horoi, Natik, Lajoie, Wolf, and
  Belilovsky]{davari2023}
MohammadReza Davari, Stefan Horoi, Amine Natik, Guillaume Lajoie, Guy Wolf, and
  Eugene Belilovsky.
\newblock Reliability of {CKA} as a similarity measure in deep learning.
\newblock In \emph{International Conference on Learning Representations}, 2023.

\bibitem[Ding et~al.(2021)Ding, Denain, and Steinhardt]{ding2021}
Frances Ding, Jean-Stanislas Denain, and Jacob Steinhardt.
\newblock Grounding representation similarity through statistical testing.
\newblock In \emph{Advances in Neural Information Processing Systems},
  volume~34, pp.\  1556--1568. Curran Associates, Inc., 2021.

\bibitem[Geirhos et~al.(2020)Geirhos, Meding, and Wichmann]{geirhos2020}
Robert Geirhos, Kristof Meding, and Felix~A. Wichmann.
\newblock Beyond accuracy: Quantifying trial-by-trial behaviour of {CNN}s and
  humans by measuring error consistency.
\newblock In \emph{Advances in Neural Information Processing Systems},
  volume~33, pp.\  13890--13902. Curran Associates, Inc., 2020.

\bibitem[Hewitt \& Liang(2019)Hewitt and Liang]{hewitt2019}
John Hewitt and Percy Liang.
\newblock Designing and interpreting probes with control tasks.
\newblock In \emph{Proceedings of the 2019 Conference on Empirical Methods in
  Natural Language Processing and the 9th International Joint Conference on
  Natural Language Processing (EMNLP-IJCNLP)}, pp.\  2733--2743, Hong Kong,
  China, 2019. Association for Computational Linguistics.
\newblock \doi{10.18653/v1/D19-1275}.

\bibitem[Khaligh-Razavi \& Kriegeskorte(2014)Khaligh-Razavi and
  Kriegeskorte]{khalighrazavi2014}
Seyed-Mahdi Khaligh-Razavi and Nikolaus Kriegeskorte.
\newblock Deep supervised, but not unsupervised, models may explain {IT}
  cortical representation.
\newblock \emph{PLOS Computational Biology}, 10\penalty0 (11):\penalty0
  e1003915, 2014.
\newblock \doi{10.1371/journal.pcbi.1003915}.

\bibitem[Klabunde et~al.(2025)Klabunde, Schumacher, Strohmaier, and
  Lemmerich]{klabunde2025}
Max Klabunde, Tobias Schumacher, Markus Strohmaier, and Florian Lemmerich.
\newblock Similarity of neural network models: A survey of functional and
  representational measures.
\newblock \emph{ACM Computing Surveys}, 57\penalty0 (9):\penalty0 1--52, 2025.
\newblock \doi{10.1145/3728458}.

\bibitem[Kornblith et~al.(2019)Kornblith, Norouzi, Lee, and
  Hinton]{kornblith2019}
Simon Kornblith, Mohammad Norouzi, Honglak Lee, and Geoffrey Hinton.
\newblock Similarity of neural network representations revisited.
\newblock In \emph{Proceedings of the 36th International Conference on Machine
  Learning}, volume~97 of \emph{Proceedings of Machine Learning Research},
  2019.

\bibitem[Kriegeskorte et~al.(2008)Kriegeskorte, Mur, and
  Bandettini]{kriegeskorte2008}
Nikolaus Kriegeskorte, Marieke Mur, and Peter~A. Bandettini.
\newblock Representational similarity analysis --- connecting the branches of
  systems neuroscience.
\newblock \emph{Frontiers in Systems Neuroscience}, 2:\penalty0 4, 2008.
\newblock \doi{10.3389/neuro.06.004.2008}.

\bibitem[Ladefoged \& Maddieson(1996)Ladefoged and Maddieson]{ladefoged1996}
Peter Ladefoged and Ian Maddieson.
\newblock \emph{The Sounds of the World's Languages}.
\newblock Blackwell, Oxford, 1996.

\bibitem[McAuliffe et~al.(2017)McAuliffe, Socolof, Mihuc, Wagner, and
  Sonderegger]{mcauliffe2017}
Michael McAuliffe, Michaela Socolof, Sarah Mihuc, Michael Wagner, and Morgan
  Sonderegger.
\newblock {Montreal Forced Aligner}: Trainable text-speech alignment using
  {Kaldi}.
\newblock In \emph{Interspeech 2017}, pp.\  498--502, 2017.
\newblock \doi{10.21437/Interspeech.2017-1386}.

\bibitem[Morcos et~al.(2018)Morcos, Raghu, and Bengio]{morcos2018}
Ari~S. Morcos, Maithra Raghu, and Samy Bengio.
\newblock Insights on representational similarity in neural networks with
  canonical correlation.
\newblock In \emph{Advances in Neural Information Processing Systems},
  volume~31, pp.\  5727--5736, 2018.

\bibitem[Nili et~al.(2014)Nili, Wingfield, Walther, Su, Marslen-Wilson, and
  Kriegeskorte]{nili2014}
Hamed Nili, Cai Wingfield, Alexander Walther, Li~Su, William Marslen-Wilson,
  and Nikolaus Kriegeskorte.
\newblock A toolbox for representational similarity analysis.
\newblock \emph{PLOS Computational Biology}, 10\penalty0 (4):\penalty0
  e1003553, 2014.
\newblock \doi{10.1371/journal.pcbi.1003553}.

\bibitem[Nycz \& Hall-Lew(2014)Nycz and Hall-Lew]{nycz2014}
Jennifer Nycz and Lauren Hall-Lew.
\newblock Best practices in measuring vowel merger.
\newblock In \emph{Proceedings of Meetings on Acoustics}, pp.\  060008, 2014.
\newblock \doi{10.1121/1.4894063}.

\bibitem[Peterson \& Barney(1952)Peterson and Barney]{petersonbarney1952}
Gordon~E. Peterson and Harold~L. Barney.
\newblock Control methods used in a study of the vowels.
\newblock \emph{The Journal of the Acoustical Society of America}, 24\penalty0
  (2):\penalty0 175--184, 1952.
\newblock \doi{10.1121/1.1906875}.

\bibitem[Raghu et~al.(2017)Raghu, Gilmer, Yosinski, and
  Sohl-Dickstein]{raghu2017}
Maithra Raghu, Justin Gilmer, Jason Yosinski, and Jascha Sohl-Dickstein.
\newblock {SVCCA}: Singular vector canonical correlation analysis for deep
  learning dynamics and interpretability.
\newblock In \emph{Advances in Neural Information Processing Systems},
  volume~30, pp.\  6076--6085, 2017.

\bibitem[Stafford et~al.(2026)Stafford, Lahaussois, and
  Wisniewski]{stafford2026}
Eli Stafford, Aim\'{e}e Lahaussois, and Guillaume Wisniewski.
\newblock What do neural speech models know about phonology? {E}vidence from
  structured phoneme confusions.
\newblock In \emph{Findings of the Association for Computational Linguistics:
  ACL 2026}, pp.\  25860--25873, San Diego, California, United States, 2026.
  Association for Computational Linguistics.
\newblock \doi{10.18653/v1/2026.findings-acl.1288}.

\bibitem[Tanner et~al.(2015)Tanner, Morgan-Short, and Luck]{tanner2015}
Darren Tanner, Kara Morgan-Short, and Steven~J. Luck.
\newblock How inappropriate high-pass filters can produce artifactual effects
  and incorrect conclusions in {ERP} studies of language and cognition.
\newblock \emph{Psychophysiology}, 52\penalty0 (8):\penalty0 997--1009, 2015.
\newblock \doi{10.1111/psyp.12437}.

\bibitem[Wells(1982)]{wells1982}
J.~C. Wells.
\newblock \emph{Accents of English}.
\newblock Cambridge University Press, Cambridge, 1982.
\newblock Three volumes.

\bibitem[Williams et~al.(2021)Williams, Kunz, Kornblith, and
  Linderman]{williams2021}
Alex~H. Williams, Erin Kunz, Simon Kornblith, and Scott~W. Linderman.
\newblock Generalized shape metrics on neural representations.
\newblock In \emph{Advances in Neural Information Processing Systems},
  volume~34, pp.\  4738--4750, 2021.

\end{thebibliography}

\appendix

\section{The two measures stated in full}
\label{app:measuredefs}

Equation~\ref{eq:rsa} of \S\ref{sec:background} defines RSA, which is the
default throughout. The other two are stated here. Linear CKA
\citep{kornblith2019} compares the Gram matrices of inner products between
conditions,
\begin{equation}
\mathbf{K} = \mathbf{X}\mathbf{X}^{\!\top}, \quad
\mathbf{L} = \mathbf{Y}\mathbf{Y}^{\!\top}, \qquad
\mathrm{CKA}(\mathbf{X},\mathbf{Y}) =
\frac{\mathrm{HSIC}(\mathbf{K},\mathbf{L})}
     {\sqrt{\mathrm{HSIC}(\mathbf{K},\mathbf{K})\,
            \mathrm{HSIC}(\mathbf{L},\mathbf{L})}},
\label{eq:cka}
\end{equation}
with $\mathrm{HSIC}(\mathbf{K},\mathbf{L}) =
\operatorname{tr}(\mathbf{K}\mathbf{H}\mathbf{L}\mathbf{H})/(N-1)^2$ and
$\mathbf{H} = \mathbf{I} - \tfrac{1}{N}\mathbf{1}\mathbf{1}^{\!\top}$ the
centring matrix, so that HSIC measures the alignment of the two structures once
the mean has been removed from each. Orthogonal Procrustes
\citep{williams2021} minimises
$\lVert \mathbf{X}\mathbf{Q} - \mathbf{Y} \rVert_F$ over rotations $\mathbf{Q}$,
which requires the two systems to be brought to a common dimension first; we
project both to a matched rank $d$, and $d$ cannot exceed $k-1$ for $k$ centred
conditions, which is why \S\ref{sec:measures} treats the rank as a bifurcation
in its own right rather than as a preprocessing detail.

\section{Experimental protocol}
\label{app:protocol}

\paragraph{Aggregation.} On the representation side, a token's interval selects
a span of frame-level representations, mean-pooled into one vector per token;
tokens carrying the same phone label are then summarised by their mean. On the
reference side, categories are summarised by the coordinate-wise median of their
tokens' $(F_1, F_2)$ pairs. Formant estimation carries occasional gross errors,
octave jumps and $F_1$--$F_2$ confusions, to which a mean is not robust, whereas
the representations show no comparably structured outliers; measuring vowel
differences by Euclidean distance between category centroids is in any case an
established approach in phonetics \citep{nycz2014}. Taking the median per
coordinate commutes with any monotone per-coordinate transformation, so the
order of the median, the Bark warping and the standardisation is immaterial: a
candidate degree of freedom that turns out to be inert, which we note because so
few of the others are. It is not equivariant to rotation of the $(F_1, F_2)$
plane, which is acceptable because those axes are interpretable rather than
arbitrary.

\paragraph{The space of analyses explored.} Not every grouping of vowel
categories would be defensible, and we do not claim otherwise: pooling two
categories at opposite corners of the plane would produce a centroid
corresponding to no attested vowel. Our principal manipulation is selection
rather than arbitrary merging. We draw subsets of the categories a language
actually distinguishes, so that every condition in every draw is a genuine
category of that language; where merging is at issue (\S\ref{sec:composition})
the draws are matched to the profile of category sizes of the published
inventory, and restricted to the dispersion spanned by attested inventories. The
space we explore is the space of defensible analyses, not the space of all
partitions.

\paragraph{Why vowels admit a reference at all.} The vowel triangle is a
measurement tool rather than a theoretical idealisation: it is reproducible in
any laboratory from a microphone and an hour of recordings, and recovered
directly from the signal, with no model and no annotation anywhere in the
procedure. This is what makes it independent of the systems
compared, rather than one more estimate from a related pipeline.

\paragraph{Two remarks on the noise ceiling.} The Spanish ceiling is exactly
\CeilMaxExact: with five categories there are only ten distances to order and
both halves ordered them identically, which is a property of a coarse grid and
not a measurement, and we use it as a denominator nowhere. The limitation that
matters more, that a split-half reliability measures reliability and not
validity, is stated in \S\ref{sec:instrument} because the recommendation of
Appendix~\ref{app:checklist} depends on it.

\paragraph{Reading the tables.} Unless a table says otherwise, its cells are
computed in the default configuration of \S\ref{sec:instrument}, with the layer
fixed at the RSA argmax; a score for a representation is the median of its
per-language scores over the fifteen languages, and an interval is a percentile
bootstrap over languages with two thousand resamples. On the anchored scale, a
measure's floor is the median of the nine random-weight controls under that
measure and its ceiling is the reference compared against itself, so each
measure has an interval of its own. Where a projection rank is varied, cells
that cannot exist are left empty rather than set to zero: a rank of $d$ requires
$k \geq d+1$ conditions, so rank and cardinality cannot be crossed freely and the
space of pipelines is not a hypercube. We are not aware of this being noted
elsewhere. Where a row corresponds to a single draw, such as the
full inventory, it has no dispersion to report. Where a table correlates two rankings of the
languages, \emph{by level} ranks them by the median score over representations
and \emph{by gap} by the gap between a self-supervised model and a spectral
baseline, each taken at both extremes of the choice. A dash marks a cell that
was not computed, which is not the same as one that cannot exist. Departures
from any of this are stated in the subsection that uses the table.

\paragraph{Units.} Frequency in hertz is not perceptually uniform: the same
interval spans a large perceptual distance low in the spectrum and a small one
high up, so Euclidean distances computed in hertz would exaggerate differences
along $F_1$ relative to $F_2$ and misrepresent the geometry the reference is
meant to encode. Warping to a perceptual scale before measuring distances is
standard practice in phonetics, and Bark is one of the scales in common use.
Each coordinate is then standardised within language.

\section{The condition set: coverage, plausible inventories, percentiles}
\label{app:composition}

The three tables below support \S\ref{sec:composition}. The first regresses the
gap on four coverage statistics of the condition set, pooled and by cardinality.
The second restricts the draws to the range of dispersion the fifteen attested
inventories span. The third places the hand-built English inventory among the
subsets and the size-matched merges it could have been. Its contrast row is
that English percentile minus the mean percentile of the four languages whose
inventory changed without one, each among random subsets of its former size,
with a bootstrap interval over both; the range printed for those four is their
spread, not the mean the contrast subtracts.

\begin{table}[t]
\centering
\small
\caption{Variance of the gap explained by four coverage statistics of the condition set, and by~its~\mbox{cardinality}~alone.}
\label{tab:coverage-rsq}
\begin{tabular}{lrrrr}
\toprule
scope & $R^2$ & raw $\sigma$ & residual $\sigma$ & $n$ \\
\midrule
k=5 & 0.096 & 0.144 & 0.138 & 500 \\
k=7 & 0.125 & 0.082 & 0.077 & 500 \\
k=12 & 0.260 & 0.047 & 0.041 & 500 \\
k=15 & 0.238 & 0.037 & 0.033 & 500 \\
k=20 & 0.338 & 0.023 & 0.019 & 500 \\
pooled: coverage only & 0.101 & 0.080 & 0.076 & 2500 \\
pooled: cardinality only & 0.000 & 0.080 & 0.080 & 2500 \\
pooled: coverage and cardinality & 0.115 & 0.080 & 0.075 & 2500 \\
\bottomrule
\end{tabular}
\end{table}

\begin{table}[t]
\centering
\small
\caption{Dispersion restricted to the range of coverage spanned by the attested inventories.}
\label{tab:band}
\begin{tabular}{lrrrrr}
\toprule
$k$ & in band (\%) & $\sigma$ all & $\sigma$ in band & reduction (\%) & mean gap \\
\midrule
5 & 67.6 & 0.144 & 0.113 & 22 & 0.036 \\
7 & 71.8 & 0.082 & 0.069 & 16 & 0.043 \\
12 & 78.6 & 0.047 & 0.041 & 12 & 0.051 \\
15 & 85.4 & 0.037 & 0.034 & 9 & 0.053 \\
20 & 95.6 & 0.023 & 0.022 & 3 & 0.057 \\
\bottomrule
\end{tabular}
\end{table}

\begin{table}[t]
\centering
\small
\caption{Where an inventory sits in the distribution of inventories that could have been chosen instead.}
\label{tab:percentile}
\begin{tabular}{lrr}
\toprule
inventory & percentile & 95\% CI \\
\midrule
hand-built table, subsets & 71.8 & {\tiny [67.6, 75.8]} \\
hand-built table, merges & 75.2 & {\tiny [71.4, 79.0]} \\
inventories changed without one (4 languages) & 38--47 & --- \\
\textbf{contrast} & +27.7 & {\tiny [+23.2, +32.1]} \\
\bottomrule
\end{tabular}
\end{table}

%

\section{Dispersion is not displacement}
\label{app:displacement}

Each effect in \S\ref{sec:results} was first reported as the quantity natural
to its analysis. Brought to one unit, the displacement of the SSL--spectral gap
in percent of the anchored interval, two of them change rank.
Table~\ref{tab:displacement} gives both for each choice.

\begin{table}[h]
\centering
\small
\caption{The quantity each choice was first reported by, against the
displacement of the SSL--spectral gap it produces. The value column mixes units
by construction: that is what the table documents.}
\label{tab:displacement}
\begin{tabular}{lp{5.2cm}rr}
\toprule
choice & quantity first reported & value & displacement (\% interval) \\
\midrule
composition & between-draw $\sigma$ of the \texttt{hubert\_large} $-$
  \texttt{mfcc\_40} gap, $k = 5$ & \CompSigmaMax & \CloseCompIqrPct\ (IQR, $k = 14$) \\
inclusion threshold & median differential effect on the same two-model gap
  & \ThrDifferential & \CloseThrPct \\
granularity & spread across systems of the per-system shift
  & \GranDifferential & \CloseGranPct \\
distance convention & spread across models of the per-model shift
  & \MetDifferential & \CloseDistPct \\
layer selection & simulated inflation of an SSL--baseline margin
  & \LayerInflationPct & \CloseLayerPct \\
projection rank & slopes of a two-model gap against the number of pairs
  & --- & \CloseRankPct \\
aggregation & spread across models of the per-model shift
  & \AggDifferentialPct & \CloseAggPct \\
\bottomrule
\end{tabular}
\end{table}

\section{The ratio, and the conventions of our own pipeline}
\label{app:tables}

Table~\ref{tab:ratio-sensitivity} gives the ratio of between-draw dispersion to
the effect of cardinality under the two choices that were not declared when it
was first computed, the distance convention of the draws and the summary of
their dispersion. Its denominator is not distinguishable from zero, which is why
\S\ref{sec:composition} reports the two quantities separately.
Table~\ref{tab:displacements} lists what a change of distance convention
displaces in our own pipeline, from a numeric shift to a change of verdict.

The reversals of verdict in Table~\ref{tab:rank-verdicts} do not come from our
pipeline. Under the change of distance convention, the slope of RSA against the
size of the condition set, at rank~2 on the full domain, changes by a factor of
\MetRsaFactor, while that of CKA is identical under both conventions, at
\MetCkaSlope: the representation-side distance is free for RSA and fixed by the
linear kernel for CKA.

\begin{table}[t]
\centering
\small
\caption{RSA minus CKA slope, paired, by projection rank, domain and distance convention.}
\label{tab:rank-verdicts}
\begin{tabular}{llllrl}
\toprule
rank & domain & distance & difference & 95\% CI & excludes 0 \\
\midrule
2 & full & euclidean & -0.0004 & {\tiny [-0.0027, +0.0020]} & no \\
2 & $k \geq 11$ & euclidean & +0.0013 & {\tiny [-0.0014, +0.0039]} & no \\
4 & full & euclidean & +0.0013 & {\tiny [-0.0010, +0.0035]} & no \\
4 & $k \geq 11$ & euclidean & +0.0083 & {\tiny [+0.0066, +0.0102]} & yes \\
10 & $k \geq 11$ & euclidean & -0.0049 & {\tiny [-0.0068, -0.0030]} & yes \\
2 & full & cosine & +0.0097 & {\tiny [+0.0062, +0.0130]} & yes \\
2 & $k \geq 11$ & cosine & +0.0142 & {\tiny [+0.0106, +0.0181]} & yes \\
4 & full & cosine & +0.0010 & {\tiny [-0.0021, +0.0038]} & no \\
4 & $k \geq 11$ & cosine & -0.0009 & {\tiny [-0.0033, +0.0013]} & no \\
10 & $k \geq 11$ & cosine & -0.0009 & {\tiny [-0.0029, +0.0010]} & no \\
\bottomrule
\end{tabular}
\end{table}

\begin{table}[t]
\centering
\small
\caption{Ratio of between-draw dispersion to the effect of cardinality, under both undeclared choices.}
\label{tab:ratio-sensitivity}
\begin{tabular}{lrrrr}
\toprule
distance & summary & $\sigma$ & range & ratio \\
\midrule
Euclidean & mean & 0.080822 & 0.004624 & 17.5 \\
cosine & mean & 0.066793 & 0.004768 & 14.0 \\
Euclidean & median & 0.061235 & 0.004624 & 13.2 \\
cosine & median & 0.047017 & 0.004768 & 9.9 \\
\bottomrule
\end{tabular}
\end{table}

\begin{table}[t]
\centering
\small
\caption{Quantities displaced by a change of distance convention, measured in our own pipeline.}
\label{tab:displacements}
\begin{tabular}{llc}
\toprule
quantity displaced & magnitude & nature \\
\midrule
raw formants vs identity control & 1.61 units & numeric \\
SSL--spectral gap & $\times$1.77 & numeric \\
percentile of the published inventory & 18.6 points & numeric \\
pooled coverage $R^2$ & $\times$4.8 & numeric \\
\textbf{RSA-vs-CKA verdict} & \textbf{4 of 5 cells} & \textbf{qualitative} \\
\bottomrule
\end{tabular}
\end{table}

Table~\ref{tab:sigma-by-k} separates what the size of the condition set does
from what its composition does (\S\ref{sec:composition}), and
Table~\ref{tab:aggregation} gives the displacement of each anchor when the two
aggregation estimators are matched (\S\ref{sec:aggregation}).
\begin{table}[t]
\centering
\small
\caption{Mean gap between two representations, and its dispersion across draws of which categories to include, by condition-set size.}
\label{tab:sigma-by-k}
\begin{tabular}{lrrr}
\toprule
$k$ & mean gap & $\sigma$ across draws & SE of mean \\
\midrule
5 & 0.055 & 0.144 & 0.0065 \\
7 & 0.054 & 0.082 & 0.0037 \\
12 & 0.059 & 0.047 & 0.0021 \\
15 & 0.058 & 0.037 & 0.0017 \\
20 & 0.058 & 0.023 & 0.0010 \\
28 & 0.059 & --- & --- \\
\bottomrule
\end{tabular}
\end{table}

\begin{table}[t]
\centering
\small
\caption{Displacement of each anchor of Table~\ref{tab:scale} when the representation side is pooled with a coordinate-wise median instead of a mean.}
\label{tab:aggregation}
\begin{tabular}{lrrr}
\toprule
family & anchor & displacement & \% interval \\
\midrule
pre-trained speech and audio models & 0.9003 & -0.0074 & -1.49 \\
non-learned spectral features & 0.8535 & -0.0114 & -2.29 \\
learned compression & 0.8421 & -0.0019 & -0.37 \\
raw formant measurements & 0.7571 & -0.0160 & -3.22 \\
architecture alone & \ScaleRandom & -0.0009 & -0.18 \\
out of modality & 0.3298 & -0.0108 & -2.16 \\
\bottomrule
\end{tabular}
\end{table}

Table~\ref{tab:closing} is the closing comparison of \S\ref{sec:hierarchy},
read by Figure~\ref{fig:bifurcations} in the body.
\begin{table}[t]
\centering
\small
\caption{Every choice on one unit: the displacement of the gap between pre-trained speech and audio models and non-learned spectral features, in percent of the anchored interval.}
\label{tab:closing}
\resizebox{\textwidth}{!}{%
\begin{tabular}{lllrrr}
\toprule
choice & extremes & $n$ & displacement & \% interval & 95\% interval \\
\midrule
composition (IQR, $k = 14$) & 25th vs 75th percentile draw & 1 language, 500 draws & \CloseCompIqrShift & \CloseCompIqrPct & {\tiny \CloseCompIqrPctCI} \\
composition ($\sigma$, $k = 14$) & between-draw dispersion & 1 language, 500 draws & \CloseCompSigmaShift & \CloseCompSigmaPct & {\tiny \CloseCompSigmaPctCI} \\
inclusion threshold & 0 ms vs 100 ms & 15 languages & \CloseThrShift & \CloseThrPct & {\tiny \CloseThrPctCI} \\
granularity & 28 categories vs 12 categories & 1 language, 1000 token resamples & \CloseGranShift & \CloseGranPct & {\tiny \CloseGranPctCI} \\
distance convention & Euclidean reference vs cosine reference & 15 languages & \CloseDistShift & \CloseDistPct & {\tiny \CloseDistPctCI} \\
layer selection & argmax vs relative depth 0.05 & 15 languages & \CloseLayerShift & \CloseLayerPct & {\tiny \CloseLayerPctCI} \\
projection rank & rank 2 vs rank 10 & 11 languages & \CloseRankShift & \CloseRankPct & {\tiny \CloseRankPctCI} \\
aggregation & mean vs median & 15 languages & \CloseAggShift & \CloseAggPct & {\tiny \CloseAggPctCI} \\
\bottomrule
\end{tabular}
}
\end{table}

\section{Alternatives to structural comparison}
\label{app:alternatives}

Two common alternatives show why a reference-based framing differs from what is
usually done; neither compares structure. Fitting a map, as in linear probing
\citep{alain2016} or the encoding models of neuroscience, makes the score depend
on the capacity of the class the map is drawn from, so that with a rich enough
class everything becomes predictable from everything; what is measured is
decodability, which is why probes are now paired with control tasks that bound
what a probe can learn unaided \citep{hewitt2019}. Comparing systems by their
behaviour is silent about representation: error consistency stays low between
systems whose accuracies match closely \citep{geirhos2020}, and systems with
very different internal geometries can produce similar error profiles
\citep{stafford2026}.

The control tasks of \citet{hewitt2019} are the nearest relative in that
literature of what \S\ref{sec:instrument} does, in that both bound a score by
something the comparison does not supply. They differ in what the bound is
about: a control task bounds what a probe can learn from a representation,
whereas the reference geometry here states what the conditions ought to look
like, so a choice that changes the answer can be shown to make it wrong about
that claim rather than merely unstable.

\section{Languages}
\label{app:languages}

\begin{table}[t]
\centering
\small
\caption{The fifteen languages, ordered by the size of their inventory.}
\label{tab:app_languages}
\resizebox{\textwidth}{!}{%
\begin{tabular}{llllrrrr}
\toprule
language & FLEURS code & family & tone (WALS 13A) & $k$ & vowel tokens & split-half ceiling & grid step \\
\midrule
Spanish & \texttt{es\_419} & Romance & none & 5 & 2,500 & 1.000 & \QuantGridEs \\
Mandarin & \texttt{cmn\_hans\_cn} & Sinitic & complex & 7 & 3,500 & 0.998 & 1.30e-03 \\
Polish & \texttt{pl\_pl} & Slavic & none & 8 & 4,000 & 0.997 & 5.47e-04 \\
Swahili & \texttt{sw\_ke} & Bantu & none & 10 & 4,063 & 0.996 & 1.32e-04 \\
Japanese & \texttt{ja\_jp} & Japonic & simple & 12 & 5,765 & 0.996 & 4.17e-05 \\
Hausa & \texttt{ha\_ng} & Chadic & simple & 13 & 6,153 & 0.997 & 2.53e-05 \\
Korean & \texttt{ko\_kr} & Koreanic & none & 13 & 6,479 & 0.997 & 2.53e-05 \\
Portuguese & \texttt{pt\_br} & Romance & not coded & 14 & 7,000 & 0.999 & 1.59e-05 \\
Russian & \texttt{ru\_ru} & Slavic & none & 14 & 6,684 & 0.995 & 1.59e-05 \\
French & \texttt{fr\_fr} & Romance & none & 15 & 7,444 & 0.998 & 1.04e-05 \\
German & \texttt{de\_de} & Germanic & none & 16 & 8,000 & 0.999 & 6.94e-06 \\
Vietnamese & \texttt{vi\_vn} & Austroasiatic & complex & 17 & 8,025 & 0.998 & 4.77e-06 \\
Thai & \texttt{th\_th} & Tai-Kadai & complex & 18 & 8,636 & 0.999 & 3.35e-06 \\
Turkish & \texttt{tr\_tr} & Turkic & none & 20 & 7,491 & 0.989 & 1.75e-06 \\
English & \texttt{en\_us} & Germanic & none & 28 & 11,718 & 0.968 & 2.22e-07 \\
\bottomrule
\end{tabular}
}
\end{table}

Fifteen of the 102 languages of FLEURS were used. They were chosen to vary along
the three dimensions we expected to bear on the geometry, the size of the vowel
inventory, the presence of lexical tone or pitch accent, and language family,
rather than by resource availability, since the size of the condition set is
itself one of the quantities under study. The resulting sample spans eleven
families, includes three languages with complex tone systems and two with
simple ones, and ranges
from five vowel categories to twenty-eight. FLEURS itself covers 102 languages
with roughly twelve hours of speech per language, recorded by three native
speakers each from a common set of sentences; its content is parallel across
languages, a property few multilingual corpora have and one that matters here,
since the language is the unit of replication.


\section{Representations}
\label{app:models}

\begin{table}[t]
\centering
\small
\caption{The twenty-seven representations, grouped by the families of Table~\ref{tab:scale}.}
\label{tab:app_models}
\resizebox{\textwidth}{!}{%
\begin{tabular}{lllrrrl}
\toprule
representation & checkpoint & revision & layers & width & layer used &  \\
\midrule
\multicolumn{7}{l}{\textit{pre-trained speech and audio models (8)}} \\
\texttt{hubert\_large} & \texttt{facebook/hubert-large-ll60k} & \texttt{ff022d0}$^{\dagger}$ & 25 & 1024 & 5 (0--24) &  \\
\texttt{mert\_330m} & \texttt{m-a-p/MERT-v1-330M} & \texttt{5240c27}$^{\dagger}$ & 25 & 1024 & 8 (1--22) &  \\
\texttt{mms\_300m} & \texttt{facebook/mms-300m} & \texttt{4ee317c}$^{\dagger}$ & 25 & 1024 & 2 (0--18) &  \\
\texttt{wav2vec2\_base} & \texttt{facebook/wav2vec2-base} & \texttt{0b5b8e8}$^{\dagger}$ & 13 & 768 & 1 (0--12) &  \\
\texttt{wav2vec2\_english} & \texttt{facebook/wav2vec2-base-960h} & \texttt{22aad52} & 13 & 768 & 0 (0--9) &  \\
\texttt{wavlm\_large} & \texttt{microsoft/wavlm-large} & \texttt{c1423ed}$^{\dagger}$ & 25 & 1024 & 3 (0--24) &  \\
\texttt{whisper\_medium} & \texttt{openai/whisper-medium} & \texttt{abdf7c3} & 25 & 1024 & 2 (0--5) &  \\
\texttt{xlsr53} & \texttt{facebook/wav2vec2-large-xlsr-53} & \texttt{c3f9d88}$^{\dagger}$ & 25 & 1024 & 2 (0--17) &  \\
\midrule \multicolumn{7}{l}{\textit{learned compression (4)}} \\
\texttt{dac\_encoder\_24khz} & DAC 24 kHz (library weights) & --- & 1 & 1024 & 0 &  \\
\texttt{encodec\_encoder\_24khz} & EnCodec 24 kHz, 6 kbps (library weights) & --- & 1 & 128 & 0 &  \\
\texttt{mimi\_encoder\_24khz} & \texttt{kyutai/mimi} & \texttt{89091b3} & 1 & 512 & 0 &  \\
\texttt{speechtokenizer\_encoder} & \texttt{fnlp/SpeechTokenizer} & \texttt{4d54939} & 1 & 1024 & 0 &  \\
\midrule \multicolumn{7}{l}{\textit{non-learned spectral features (3)}} \\
\texttt{logmel\_80} & no training & --- & 1 & 80 & 0 &  \\
\texttt{mfcc13\_delta} & no training & --- & 1 & 39 & 0 &  \\
\texttt{mfcc\_40} & no training & --- & 1 & 40 & 0 &  \\
\midrule \multicolumn{7}{l}{\textit{raw formant measurements (1)}} \\
\texttt{formants\_f1f4} & F1--F4, formant tracking (no training) & --- & 1 & 4 & 0 &  \\
\midrule \multicolumn{7}{l}{\textit{architecture alone (9)}} \\
\texttt{hubert\_large\_random\_seed0} & \texttt{facebook/hubert-large-ll60k} & \texttt{ff022d0}$^{\dagger}$ & 25 & 1024 & 0 (0--18) & seed 0 \\
\texttt{hubert\_large\_random\_seed1} & \texttt{facebook/hubert-large-ll60k} & \texttt{ff022d0}$^{\dagger}$ & 25 & 1024 & 5 (0--24) & seed 1 \\
\texttt{hubert\_large\_random\_seed2} & \texttt{facebook/hubert-large-ll60k} & \texttt{ff022d0}$^{\dagger}$ & 25 & 1024 & 1 (0--21) & seed 2 \\
\texttt{wav2vec2\_base\_random\_seed0} & \texttt{facebook/wav2vec2-base} & \texttt{0b5b8e8}$^{\dagger}$ & 13 & 768 & 0 (0--12) & seed 0 \\
\texttt{wav2vec2\_base\_random\_seed1} & \texttt{facebook/wav2vec2-base} & \texttt{0b5b8e8}$^{\dagger}$ & 13 & 768 & 0 (0--12) & seed 1 \\
\texttt{wav2vec2\_base\_random\_seed2} & \texttt{facebook/wav2vec2-base} & \texttt{0b5b8e8}$^{\dagger}$ & 13 & 768 & 0 (0--9) & seed 2 \\
\texttt{xlsr53\_random\_seed0} & \texttt{facebook/wav2vec2-large-xlsr-53} & \texttt{c3f9d88}$^{\dagger}$ & 25 & 1024 & 3 (0--24) & seed 0 \\
\texttt{xlsr53\_random\_seed1} & \texttt{facebook/wav2vec2-large-xlsr-53} & \texttt{c3f9d88}$^{\dagger}$ & 25 & 1024 & 2 (0--18) & seed 1 \\
\texttt{xlsr53\_random\_seed2} & \texttt{facebook/wav2vec2-large-xlsr-53} & \texttt{c3f9d88}$^{\dagger}$ & 25 & 1024 & 1 (0--14) & seed 2 \\
\midrule \multicolumn{7}{l}{\textit{out of modality (2)}} \\
\texttt{vit\_imagenet\_10hz} & \texttt{google/vit-base-patch16-224-in21k} & \texttt{b456956} & 13 & 768 & 6 (1--10) & 10 Hz \\
\texttt{vit\_imagenet\_50hz} & \texttt{google/vit-base-patch16-224-in21k} & \texttt{b456956} & 13 & 768 & 6 (1--12) & 50 Hz \\
\bottomrule
\end{tabular}
}
\end{table}

A dagger on a revision marks a checkpoint for which the extraction cache holds
two revisions and the pipeline did not record which was used
(Appendix~\ref{app:caught}); a control inherits the dagger of the trained model
whose configuration it reads.

Every representation is evaluated at the layer selected as described in
\S\ref{sec:instrument}; \S\ref{sec:layer} reports what that selection is worth.

\paragraph{Random-weight controls.} The nine controls are \CtrlArchitectures{}
architectures, those of \texttt{hubert\_large}, \texttt{wav2vec2\_base} and
\texttt{xlsr53}, each instantiated under seeds \CtrlSeeds. A control reads the
configuration of its trained checkpoint, sets the seed with
\texttt{torch.manual\_seed} and builds the model from that configuration alone,
through \texttt{AutoModel.from\_config} of \texttt{transformers}
\CtrlTransformers{} (PyTorch \CtrlTorch); no trained weight is loaded, the
convolutional feature encoder included. The weights therefore follow the
library's default initialisation for that class: linear layers drawn from a
centred normal of standard deviation \CtrlInitStd{} (the checkpoint's
\texttt{initializer\_range}) with zero biases, normalisation layers set to the
identity, and convolutions drawn from a He normal; the wav2vec~2.0 class
additionally gives its positional convolution and its feature projection laws
scaled by their fan-in. Layer count and width are those of the trained model
(Table~\ref{tab:app_models}), so each control differs from its trained
counterpart by training alone.
A control is scored like any other
representation: for each language, at its best layer, then the median over the
fifteen languages. The floor of the anchored scale is the median of these nine
scores. It pools architectures and seeds: it says where an untrained network of
this family falls, not how much one architecture varies from seed to seed.

\paragraph{Vision checkpoint.} The out-of-modality floor is one checkpoint,
\texttt{google/vit-base-patch16-224-in21k}, at the revision given in
Table~\ref{tab:app_models}: a ViT-Base with patches of \VitPatch{} pixels on
\VitImage{}-pixel images, pre-trained on ImageNet-21k, as its identifier
states, and never exposed to audio. Audio reaches it as images. The signal,
resampled to \VitSampleRate{}~Hz, gives one log-mel spectrogram (\VitMels{}
bands, an FFT of \VitFft{} samples, a hop of \VitMelHopMs{}~ms, decibels
relative to the utterance maximum); the spectrogram is cut into windows of
\VitWindowMs{}~ms, each window is rescaled to $[0,1]$ by its own minimum and
maximum, resized bilinearly to \VitImage{}$\times$\VitImage{} and copied to the
three colour channels, without the ImageNet normalisation. One window gives one
frame: the class token of each of the \VitLayers{} hidden states (the patch
embedding and each transformer block), of width \VitWidth. The two rows of
Table~\ref{tab:app_models} are two strides between windows, named
\VitNamedFifty{}~Hz and \VitNamedTen{}~Hz in the configuration. Because windows
start on whole mel frames, the first advances by one mel frame
(\VitMelHopMs{}~ms, \VitRateFifty{} frames per second) and the second by
\VitHopTenMs{}~ms (\VitRateTen{} frames per second); the names are kept as
labels. A vowel is represented, as for every other model, by the mean of the
frames that fall in its interval, and the layer is selected as for every other
model. The two strides give two sets of frames, hence two scores and two rows.
That there are two is a matter of record rather than of design: the coarser
stride was introduced as a speed measure, raising the configured hop from 20 to
100~ms so as to cut the number of windows fivefold, and the two were then kept
side by side deliberately, as an ablation on whether temporal resolution
changes what is recovered of the vowel geometry.

Two properties of that construction are worth stating plainly, because they
bound what the last column of Table~\ref{tab:app_models} may be used for. A
window is labelled by the mel frame it starts on rather than by its centre, and
the frame rate is recovered after the fact, as the number of frames divided by
the duration of the utterance; the correspondence between a time in the signal
and a frame index therefore drifts along the utterance rather than holding
exactly. That column is accordingly descriptive. It names a configuration and
enters no denominator anywhere in this paper -- in particular not the anchored
interval of \S\ref{sec:instrument}, which is computed from scores and never
from frame rates -- and no quantity reported in the body is derived from it.


\section{Why no inventory is canonical}
\label{app:inventories}

\S\ref{sec:intro} claims that how many degrees of vowel height to recognise, or
whether a diphthong is one condition or two, are contested questions rather than
matters of presentation. This appendix makes that claim good, states what our
corpus supplied and what we did with it, and gives the inventory used for every
language.

\paragraph{The count is an analytic decision, not an observation.}
A vowel inventory is not read off a signal. It is the output of decisions about
which contrasts are phonemic, and four of those decisions recur across the
descriptive literature \citep{ladefoged1996}. Whether length is a property of
the vowel or of the syllable determines whether a long and a short vowel are one
category or two. Whether a diphthong is a single moving target or a sequence of
two determines whether it occupies one row of the matrix or two. Whether
centralised or reduced vowels are categories in their own right or allophones of
full vowels determines how many rows unstressed syllables contribute. And how
finely height is divided, into three levels or four or five, is a choice about
which the descriptive tradition of a single language can change within a
generation. Each decision is defensible either way, each changes $N$, and none of
them is a decision about the recording.

\paragraph{Two defensible inventories for the same language, in this paper.}
We do not have to look outside our own material for an example. The English
inventory of our corpus carries twenty-eight categories as transcribed, and the
merge table of \S\ref{sec:granularity} reduces it to twelve. Both describe the
same recordings. The first follows the phone set of the aligner
\citep{mcauliffe2017}; the second follows the lexical sets of
\citet{wells1982}, which group vowels by the words that carry them rather than
by a phonemic analysis, and which are the standard descriptive device for
comparing English accents whose phoneme counts differ. Neither is a coarsening
of a true inventory that the other reports correctly. \S\ref{sec:granularity}
measures what passing between them does: every score rises, and it rises most
for the weakest representations.

\paragraph{What we used, and what we built.}
Inventories are used as observed. For each language the categories are the vowel
phones the aligner's phone set distinguishes, with no merging, which is why the
sample ranges from five categories to twenty-eight without our having chosen
that range. Two exceptions are declared. The twelve-category English merge table
is ours: it was built for this project rather than borrowed, organised by the
lexical sets above, and archived with the configuration. And the draws of
\S\ref{sec:composition} are subsets of the observed categories, never arbitrary
partitions, so that every condition in every draw is a category the language
actually distinguishes (Appendix~\ref{app:protocol}).

\begin{table}[t]
\centering
\small
\caption{The inventory used for each language, as the aligner transcribes it, with no merging.}
\label{tab:app_inventories}
\resizebox{\textwidth}{!}{%
\begin{tabular}{lrp{0.72\textwidth}}
\toprule
language & $k$ & categories \\
\midrule
Spanish & 5 & a e o i u \\
Mandarin & 7 & i o a u e y \textschwa{} \\
Polish & 8 & a \textepsilon{} \textopeno{} \textbari{} i u \~{\textopeno{}} \~{\textepsilon{}} \\
Swahili & 10 & \textscripta{} i u \textepsilon{} \textopeno{} \textscripta{}\textlengthmark{} u\textlengthmark{} i\textlengthmark{} \textepsilon{}\textlengthmark{} \textopeno{}\textlengthmark{} \\
Japanese & 12 & a o i e o\textlengthmark{} \textturnm{} e\textlengthmark{} \textbari{} \textbari{}\textlengthmark{} i\textlengthmark{} a\textlengthmark{} \textturnm{}\textlengthmark{} \\
Hausa & 13 & a\textlengthmark{} a \textsci{} \textschwa{} e\textlengthmark{} o\textlengthmark{} i\textlengthmark{} \textupsilon{} \textepsilon{} u\textlengthmark{} e \textopeno{} o \\
Korean & 13 & \textturna{} i \textturnv{} o e u \textepsilon{}\textlengthmark{} \textbari{} \textturnv{}\textlengthmark{} i\textlengthmark{} o\textlengthmark{} e\textlengthmark{} u\textlengthmark{} \\
Portuguese & 14 & \textturna{} i a u \textbari{} o e \~{\textturna{}} \~{e} \textepsilon{} \textopeno{} \~{o} \~{i} \~{u} \\
Russian & 14 & a \textturna{} o e \textschwa{} i \textbari{} \textupsilon{} \textsci{} u \textepsilon{} \ae{} \textbaro{} \textbaru{} \\
French & 15 & a e i \textepsilon{} \~{\textscripta{}} \textopeno{} \~{\textopeno{}} y \~{\textepsilon{}} o u \o{} \oe{} \textscripta{} \textschwa{} \\
German & 16 & \textschwa{} a \textsci{} \textturna{} \textepsilon{} i\textlengthmark{} e\textlengthmark{} a\textlengthmark{} \textupsilon{} \textopeno{} o\textlengthmark{} u\textlengthmark{} y\textlengthmark{} \textscy{} \o{}\textlengthmark{} \oe{} \\
Vietnamese & 17 & a\textlengthmark{} i\textlengthmark{} \textschwa{}\textlengthmark{} \textopeno{}\textlengthmark{} a \textbari{} o\textlengthmark{} \textschwa{} e\textlengthmark{} \textbari{}\textlengthmark{} \textopeno{} i \textepsilon{}\textlengthmark{} u u\textlengthmark{} o e \\
Thai & 18 & a\textlengthmark{} a i\textlengthmark{} \textopeno{}\textlengthmark{} o \textepsilon{}\textlengthmark{} i u\textlengthmark{} e\textlengthmark{} o\textlengthmark{} e \textepsilon{} \textramshorns{}\textlengthmark{} \textturnm{}\textlengthmark{} u \textturnm{} \textopeno{} \textramshorns{} \\
Turkish & 20 & a e i \textturnm{} o u y \o{} \textepsilon{} \textbari{} \textupsilon{} a\textlengthmark{} \textturnm{}\textlengthmark{} u\textlengthmark{} e\textlengthmark{} o\textlengthmark{} \o{}\textlengthmark{} i\textlengthmark{} \textscy{} y\textlengthmark{} \\
English & 28 & a i \textschwa{} \textepsilon{} \textsci{} \textturnscripta{} e i\textlengthmark{} \textturna{} \textscripta{} \textturnscripta{}\textlengthmark{} \textopeno{} \ae{} e\textlengthmark{} \textrhookschwa{} \textbaru{}\textlengthmark{} u\textlengthmark{} a\textlengthmark{} \textscripta{}\textlengthmark{} \textrevepsilon{}\textlengthmark{} o\textlengthmark{} \textbaru{} o \textrhookrevepsilon{} \textepsilon{}\textlengthmark{} u \textupsilon{} \textrevepsilon{} \\
\bottomrule
\end{tabular}
}
\end{table}

\paragraph{The same underdetermination outside speech.}
Nothing in the argument is particular to vowels. A comparison over ImageNet
classes must decide which classes and at what level of the hierarchy, and the
answer is not given by the images; a comparison against neural recordings must
decide which experimental conditions count as distinct, and that is the question
the experiment was designed to ask rather than one it answers. The reference
geometry is what our domain supplies and most do not
(\S\ref{sec:instrument}); the underdetermination of the condition set is what
our domain shares with them.

\section{The layer-selection artefact in vision}
\label{app:vision}

The layer-selection artefact of \S\ref{sec:layer} reproduces in a second
modality. On dSprites, whose images are generated from known factors, three
trained vision models (DINOv2, CLIP, ResNet-50) are compared with four
unlearned baselines under the three protocols of Table~\ref{tab:app_vision}.
Under the best layer per condition, a baseline contests \VisContestedArgmax\ of
the \VisPairs\ pairs; at one relative depth for every model, the rule of
\S\ref{sec:layer}, it contests \VisContestedMatched, and raw pixels are ahead of
all three trained models. The artefact is therefore more visible here than in
speech, where no baseline contests a model under either protocol. The setting
also has a property ours lacks: its reference geometry is exact rather than
measured, which would make the \emph{bias} of the estimator measurable there,
and not only its variance. The composition of the condition set was not tested
in vision, and the result of \S\ref{sec:hierarchy} remains a result about
speech.
\begin{table}[t]
\centering
\small
\caption{The layer-selection artefact in a second modality (dSprites; three factors $\times$ three values of $k$ = nine conditions).}
\label{tab:app_vision}
\resizebox{\textwidth}{!}{%
\begin{tabular}{lrrr}
\toprule
 & argmax & matched depth & leave-one-out \\
\midrule
\textit{pairs a baseline contests, of \VisPairs} & \VisContestedArgmax & \VisContestedMatched & \VisContestedLoo \\
\midrule \multicolumn{4}{l}{\textit{mean $\rho$ over the nine conditions}} \\
DINOv2-base & 0.629 & 0.448 & 0.530 \\
CLIP ViT-B/16 & 0.649 & 0.466 & 0.561 \\
ResNet-50 & 0.731 & 0.605 & 0.605 \\
raw pixels & 0.693 & 0.693 & 0.693 \\
PCA-64 on pixels & 0.568 & 0.568 & 0.568 \\
Gabor bank & -0.057 & -0.057 & -0.057 \\
HOG & 0.564 & 0.564 & 0.564 \\
\bottomrule
\end{tabular}
}
\end{table}

\section{What this paper does not establish}
\label{app:limitations}

The paragraph of \S\ref{sec:conclusion} states these in brief. Each is given
here with what would be needed to lift it.

\paragraph{The two decisions do not have the same reach.}
\S\ref{sec:composition} and the estimator of \S\ref{sec:aggregation} bear on how
inputs are grouped, and therefore on work in which a row aggregates several
inputs: the neuroscience tradition these measures come from, analyses at the
level of categories, and comparisons between a model and trial-averaged
recordings. \S\ref{sec:threshold} bears on which inputs are admitted, which
applies wherever these measures are used. A reader whose rows are individual
examples should take the second and not the first.

\paragraph{One domain, and one language for two results.}
Vowel geometry is the instrument. One result travels: the layer-selection
artefact of \S\ref{sec:layer} reproduces, more sharply, on a dataset whose
generative factors supply the reference (Appendix~\ref{app:vision}). The others
were not tested in a second modality, and whether their magnitudes transfer is
not established here; what transfers in any case is the argument that a choice
upstream of the matrix is beyond the reach of any invariance. Within the domain,
the composition and granularity results rest on English alone, the only language
of our sample with twenty-eight categories.

\paragraph{The space is not enumerated.}
Two choices that meet our criterion are not examined: the units the reference is
expressed in, and how representations are normalised. We do not claim to have
enumerated the space.

\paragraph{Two bifurcations, two sets of models.}
The duration threshold was swept over eight representations and the distance
convention over twenty-seven, so the wider interval on the first reports the
count rather than a less stable ranking. Extending the sweep is a matter of
compute and not of method, and the Reproducibility Statement says what it would
take.

\paragraph{What we did not measure.}
The instrument would also permit asking which measure best recovers a known
geometry at a fixed condition set. We point at this rather than pursue it,
because \S\ref{sec:measures} implies the answer would itself be conditional on
the choice this paper is about.

\section{The record: predictions and defects}
\label{app:caught}

A paper about choices that go unreported owes an account of its own. We
registered each prediction before computing it and kept a journal of the defects
found in our own pipeline. This appendix gives both, grouped by what would have
been needed to notice them; Appendix~\ref{app:preregistration} shows which
predictions can be checked from outside the project and which cannot.

Two things in the record bear on how the results of \S\ref{sec:results} should
be read. Half the predictions we registered were refuted outright, among them two whose
premise was correct and whose conclusion was not (\S\ref{sec:threshold},
\S\ref{sec:aggregation}); and the defects that cost most were those where the
artefact was internally consistent, since an error that produces a plausible
number and a file that is merely stale both survive every check that asks
whether the pipeline ran.

\paragraph{Predictions registered, and refuted.}

Of the ten predictions we registered, five were refuted outright, four were
confirmed only in part and one was never tested. Two refutations are worth
more than a line, because what they overturned was a way of reasoning rather
than a number.

\emph{The premise held and the conclusion did not.} We predicted that matching
the two aggregation estimators would narrow the gap, on the ground that the
spectral baselines carry the heavier-tailed within-category distributions. The
premise is correct and the gap widens (\S\ref{sec:aggregation}). A mechanism can
be right about what it describes and wrong about what follows from it.

\emph{The cheap diagnostic, twice.} Twice we predicted that a quantity available
without recomputation would predict a bifurcation's differential effect, and
twice the diagnostic ordered the systems correctly on the property it measures
and failed to predict the consequence (\S\ref{sec:threshold},
\S\ref{sec:aggregation}). The pattern is the finding, and it is what makes the
recommendation to sweep rather than diagnose worth stating.

A third case never reached a result. An early version of the unaggregated
analysis produced a correlation near $-0.9$ with a $p$-value indistinguishable
from zero, between the number of categories present in a draw and the dispersion
of its level. The quantity says nothing: dispersion is constant within a level,
so the correlation only restates that both terms move with the number of rows,
and it would have come out large whatever the mechanism. It was caught because
the protocol addendum, written three days earlier, had specified the test at
fixed $n$, which is the one the prediction engaged and which returns nothing.
This is the failure of \S\ref{sec:composition}'s ratio in another guise: a
statistic whose value is imposed by its construction rather than by the data.

One choice we declined to change. Late in the work we found that the family we
had been calling self-supervised contained two supervised encoders. Redefining
it would have been defensible and we did not, because the panel was named in the
pre-registration of the closing analyses and changing it after seeing their
results is the move this paper documents in others. We report the sensitivity
instead (\S\ref{sec:hierarchy}).

\paragraph{Four classes of defect.}

The defects we found fall into four kinds, distinguished by what would have been
needed to notice them.

\paragraph{Class A: an error that produced a plausible result.} An asymmetry in
how the two sides were pooled produced an effect four times its true size in
one language, which stood for six weeks as the motivating result of an earlier
draft. It was plausible, it was in the right direction, and nothing about the
output marked it. The entry below is of the same class, with a mechanism the
others do not share.

\paragraph{A storage type decided the order of the smallest distances.}
The out-of-modality anchor of Table~\ref{tab:scale} was printed as
\PrecAnchorBefore\ in one table and \PrecAnchorAfter\ in another. The second was
the value computed in double precision and the first the value computed in
single precision, and it took three hypotheses, two of them wrong, to establish
it. Until its last regeneration, the pipeline computed cosine distances between
category centroids held in single precision. For most representations this is
harmless. It is not for those whose centroids are nearly collinear: randomly
initialised networks, the vision model and the last layers of some trained
encoders, where the smallest cosine distances are of the order of $10^{-6}$ and
the subtraction that defines them loses most of their digits to cancellation.
There the order of the small distances, which is all a rank correlation sees,
was set by the last bits of the representation. Of the \PrecCells\
model--language--layer values, \PrecMoved\ moved by more than $10^{-5}$ once the
distances were computed in double precision, and the largest moved from
\PrecMaxBefore\ to \PrecMaxAfter, or \PrecMaxShiftPct\ of the anchored interval.
At the layers this paper reports, the displacement is at most \PrecDefaultMax,
and the default layer changed for \PrecArgmaxMoved\ model--language pair. No
conclusion changed, and the anchored interval moved in its fifth decimal.

Two features make the case worth recording beyond its correction. The loss
strikes where there is no structure: the representations it affects are those
that bound the scale from below, against which every other value is read. This
is the lattice of \S\ref{sec:measures} seen from the other end. There, the
estimator cannot resolve below the step of its grid; here, it cannot resolve
below the precision of its storage, and near either limit the order of the
distances is decided by something other than the representation. And the
natural remedy would have hidden the problem. Rounding both computations, or
recomputing in single precision to match, produces agreement, and agreement is
more plausible than disagreement: the obvious diagnostic confirms instead of
revealing.
The build now declares every artefact that carries this quantity and stops if
any two disagree beyond the rounding of storage.

\paragraph{Class B: the correction depended on what was neither in the code nor
in the output.} A switch that controlled the pooling above was read from a
configuration file that nothing wrote to, so that setting it changed nothing and
the code that consumed it was correct. A second instance is below; a third is
that the artefacts on disk predate a change to the extraction that would have
made a later analysis possible, and nothing about them says so: they load, they
carry the expected keys, and the only symptom is a downstream script refusing a
file that looks intact.

\paragraph{What identified a computation was not written down when it was made.}
Twice, a piece of information that was available at the moment of computing was
not stored with the result, and could not be recovered afterwards. The first is
the identity of the token. Our store of representations kept each vector and its
phone label and dropped what linked it to the token it came from, so that no row
could be put back beside the duration or the formants measured on the same
segment. Nothing was falsified; two analyses simply became impossible. The
duration sweep of \S\ref{sec:threshold} could not go beyond \HierModelsThrN\ of
\HierModelsMetN\ representations, and the unaggregated arm could not be run
until the representations were extracted again with their provenance. The second
is the revision of each checkpoint. Of the \ProvHubCheckpoints\ checkpoints taken
from a public hub, \ProvAmbiguous\ have two revisions in the cache of the
machine that extracted them, and nothing the pipeline kept says which one was
used (Appendix~\ref{app:models}). This prevents nothing that we report; it
weakens the claim that anyone can reproduce it. The two costs differ, one closing
an analysis for months and the other opening a doubt that cannot be closed, but
the mechanism is the same, and so is the remedy (Appendix~\ref{app:checklist}). A
variant of it met us in the vision arm: two files of the same name, at the same
path on two machines, one a partial run and the other the complete one, and
nothing in either said which was which. Like a table that has stopped being
generated but is still included, each looked right and only one was.

\paragraph{Class C: the code was correct and gave two answers.} The percentile
of \S\ref{sec:composition} moved by seven points between two runs of the same
script on the same data. Neither run was wrong. The draws were generated from a
shared random stream, and an earlier part of the analysis consumed a different
number of values in each run, so the subsets drawn afterwards differed. The
defect is not in any line of the code but in the coupling between two parts of
it that appear independent.

\paragraph{Class D: the check did not check.} Three controls ran, reported
nothing, and did not look where we believed they looked: the pooling switch
above, a test suite that collected no test file and therefore passed, and the
entry below. This is the class we would expect to be rarest and it is the
best attested.

\paragraph{A check that passed because it did not look.}
The build checks that no table is wider than the text, since an over-wide table
does not fail: it runs into the margin, and what passes the edge of the page is
not printed. The check read the compilation log and attributed each overfull box
to the file open when it was reported. Parentheses in the log that are not file
names closed the table's file early, so that its overflows were attributed to
nothing and the check passed. Meanwhile a column of the closing table ran past
the page, in a build that reported no error. The check existed, it ran, and it
did not look where we believed it looked; it now fails on any overflow it cannot
attribute.

\section{Preregistrations}
\label{app:preregistration}

Table~\ref{tab:app_prereg} gives, for each prediction of Appendix~\ref{app:caught},
the first commit that carries it and what that commit can prove. Not every
prediction can be checked from outside: two were written in a plan that is not
versioned, and the journal that records the verdicts entered version control
only after the first six had been computed. We report the discipline as it was,
imperfect, rather than as it should have been, because it still caught what
Appendix~\ref{app:caught} describes.
\begin{table}[t]
\centering
\small
\caption{The predictions of Appendix~\ref{app:caught}, with the commit that carries each before its computation, and what that commit proves.}
\label{tab:app_prereg}
\resizebox{\textwidth}{!}{%
\begin{tabular}{lp{0.34\textwidth}lp{0.2\textwidth}p{0.18\textwidth}}
\toprule
 & prediction & first commit & verifiable by & verdict \\
\midrule
P1 & The noise ceiling falls mechanically with the number of pairs & \texttt{5f2d11b}, 2026-09-08 & the script that states it, before it ran & refuted, in the opposite direction \\
P2 & RSA, CKA and Procrustes columns nearly identical on the first bifurcations & --- & no external trace & refuted \\
P3 & After normalisation, the SSL - MFCC gap separates tonal from non-tonal languages & --- & no external trace & stated; not tested \\
P4 & The Procrustes slope in k survives a rank control & \texttt{3a3221c}, 2026-09-08 & the script that states it, before it ran & refuted, sign reversed at rank 2 \\
P5 & The threshold's differential effect is largest in Russian & \texttt{3452e65}, 2026-09-08 & the script that states it, before it ran & refuted on both parts \\
P6 & The paired estimator narrows the SSL - spectral gap & \texttt{0c8db60}, 2026-09-15 & the script that states it, before it ran & refuted, the gap widens \\
P7 & Without aggregation the between-draw dispersion becomes negligible & \texttt{6354f5d}, 2026-09-18 & its preregistration file & principal confirmed, subsidiary refuted \\
P8 & Composition ranks above the distance convention on one unit & \texttt{e47871c}, 2026-09-23 & its preregistration file & confirmed; subsidiary refuted under the IQR \\
P9 & The three measures rank the representations alike & \texttt{e47871c}, 2026-09-23 & its preregistration file & confirmed over 27, partial over 16 \\
P10 & The ranking of representations survives composition & \texttt{13855b9}, 2026-09-23 & its preregistration file & confirmed at k = 14, collapses at k = 5 \\
\bottomrule
\end{tabular}
}
\end{table}

%

\section{A checklist}
\label{app:checklist}

The items below follow from what \S\ref{sec:results} showed to be consequential
and from what Appendix~\ref{app:caught} shows to be easy to miss. Several of them we
wrote after the defect they would have caught. The first three are the findings
of this paper turned into practice and are stated in \S\ref{sec:conclusion}; the
rest come from our own record.

\begin{itemize}
\item \textbf{Report the condition set, category by category.} No summary of the
  set stands in for the set: four coverage statistics capture a tenth of what is
  at issue (\S\ref{sec:composition}).
\item \textbf{Report the inclusion rule, and sweep it.} Which inputs are
  admitted take a comparison from \ThrPlBefore\ to \ThrPlAfter\ in one language
  of fifteen, and no diagnostic available without recomputation predicts what it
  will do
  (\S\ref{sec:threshold}).
\item \textbf{Compare at matched depth, or say that you did not.} Reporting each
  system at its best layer compares a maximum over many draws with a maximum
  over one, and accounts for \LayerShare\ of the margin (\S\ref{sec:layer}).
\item \textbf{Check that the estimator can resolve the comparison.} A rank
  correlation moves on a grid whose step grows as the condition set shrinks, and
  below some size the difference reported is smaller than the step
  (\S\ref{sec:measures}).
\item \textbf{Report a noise ceiling, and state what it does not cover.} A
  split-half reliability bounds what the reference can support; a systematic
  measurement error reproduces itself in both halves and leaves it untouched, so
  the control is blind to exactly the failure that would matter most
  (\S\ref{sec:instrument}).
\item \textbf{Compute distances in the precision in which they will be
  compared.} A rank correlation is decided by the order of the smallest
  distances, and the representations that bound a scale from below are those
  whose smallest distances fall under the resolution of single precision
  (Appendix~\ref{app:caught}).
\item \textbf{Record what identifies what was computed, at the moment it is
  computed.} The token behind each representation and the revision of each
  checkpoint cost nothing to store then and cannot be recovered later
  (Appendix~\ref{app:caught}).
\item \textbf{Before reporting a correlation, check that both its terms vary
  within the unit it is computed over.} A statistic whose value is imposed by
  its construction comes out large whatever the mechanism, and a ratio whose
  denominator is not distinguishable from zero has no value to report
  (\S\ref{sec:composition}).
\item \textbf{Make a control fail when it cannot check.} Three of the defects we
  found passed a control that ran and did not look (Appendix~\ref{app:caught}).
\end{itemize}

A checklist is not a guarantee, and Appendix~\ref{app:caught} is the evidence: we were
following one, and it did not contain these items until each of them was needed.

\end{document}